\documentclass[journal]{IEEEtran}
  \usepackage[pdftex]{graphicx}
\usepackage{amsmath}
\usepackage{listings}

\usepackage{booktabs}
\usepackage{comment}
\usepackage{color}
\usepackage[noadjust]{cite}

\usepackage{url}

\begin{document}
%
\title{Data Augmentation using Random Image Cropping and Patching for Deep CNNs}
%
%
\author{Ryo~Takahashi,
        Takashi~Matsubara,~\IEEEmembership{Member,~IEEE},
        and~Kuniaki~Uehara,
\thanks{R.~Takahashi, T.~Matsubara, and K.~Uehara are with Graduate School of System Informatics, Kobe University, 1-1 Rokko-dai, Nada, Kobe, Hyogo, 657-8501 Japan. E-mails: takahashi@ai.cs.kobe-u.ac.jp, matsubara@phoenix.kobe-u.ac.jp, and uehara@kobe-u.ac.jp.}
}

%
%


\markboth{Journal of \LaTeX\ Class Files,~Vol.~14, No.~8, August~2015}%
{Shell \MakeLowercase{\textit{et al.}}: Bare Demo of IEEEtran.cls for IEEE Journals}
%



\maketitle

\begin{abstract}
Deep convolutional neural networks (CNNs) have achieved remarkable results in image processing tasks.
However, their high expression ability risks overfitting.
Consequently, data augmentation techniques have been proposed to prevent overfitting while enriching datasets.
Recent CNN architectures with more parameters are rendering traditional data augmentation techniques insufficient.
In this study, we propose a new data augmentation technique called \emph{random image cropping and patching} (\emph{RICAP}) which randomly crops four images and patches them to create a new training image.
Moreover, RICAP mixes the class labels of the four images, resulting in an advantage of the soft labels.
We evaluated RICAP with current state-of-the-art CNNs (e.g., the shake-shake regularization model) by comparison with competitive data augmentation techniques such as cutout and mixup.
RICAP achieves a new state-of-the-art test error of \textcolor{red}{$2.19\%$} on CIFAR-10.
We also confirmed that deep CNNs with RICAP achieve better results on classification tasks using CIFAR-100 and ImageNet, an image-caption retrieval task using Microsoft COCO, and other computer vision tasks.
\end{abstract}


%
\IEEEpeerreviewmaketitle


\section{Introduction}
\label{sec:introroduction}
Deep convolutional neural networks (CNNs)~\cite{LeCun1989} have led to significant achievement in the fields of image classification and image processing owing to their numerous parameters and rich expression ability~\cite{Zeiler2014,Sermanet2014}.
A recent study demonstrated that the performance of CNNs is logarithmically proportional to the number of training samples~\cite{Sun2017}.
Conversely, without enough training samples, CNNs with numerous parameters have a risk of overfitting because they memorize detailed features of training images that cannot be generalized~\cite{Zeiler2014,Zintgraf2017}.
Since collecting numerous samples is prohibitively costly, data augmentation methods have been commonly used~\cite{Ciresan2011,Ciresan2012}.
Data augmentation increases the variety of images by manipulating them in several ways such as flipping, resizing, and random cropping~\cite{Krizhevsky2009,Szegedy2014,Simonyan2015a,He2016}.
Color jitter changes the brightness, contrast, and saturation, and color translating alternates intensities of RGB channels using principal component analysis (PCA)~\cite{Krizhevsky2012}.
Dropout on the input layer~\cite{Hinton2012} is a common technique that injects noise into an image by dropping pixels and a kind of data augmentations~\cite{Zhao2019a}.
Unlike conventional data augmentation techniques, dropout can disturb and mask the features of original images.
Many recent studies have proposed new CNN architectures that have many more parameters~\cite{He2016b,Zagoruyko2016,Huang2016b,Han2016,Xie2017}, and the above traditional data augmentation techniques have become insufficient.

Therefore, nowadays, new data augmentation techniques have attracted increasing attention~\cite{DeVries2017,Zhong2017a,Zhang2017}.
Cutout~\cite{DeVries2017} randomly masks a square region in an image at every training step and thus changes the apparent features.
Cutout is an extension of dropout on the input layer that can achieve better performance.
Random erasing~\cite{Zhong2017a} also masks a subregion in an image like cutout.
Unlike cutout, it randomly determines whether to mask a region as well as the size and aspect ratio of the masked region.
Mixup~\cite{Zhang2017} alpha-blends two images to form a new image, regularizing the CNN to favor simple linear behavior in-between training images.
In addition to an increase in the variety of images, mixup behaves as soft labels as it mixes the class labels of two images with the ratio $\lambda:1-\lambda$~\cite{Szegedy2016}.
These new data augmentation techniques have been applied to modern deep CNNs and have broken records, demonstrating the importance of data augmentation.

\begin{figure}[!t]{}
\centering
\includegraphics[width=3.8in,bb= 0 0 820 580,clip]{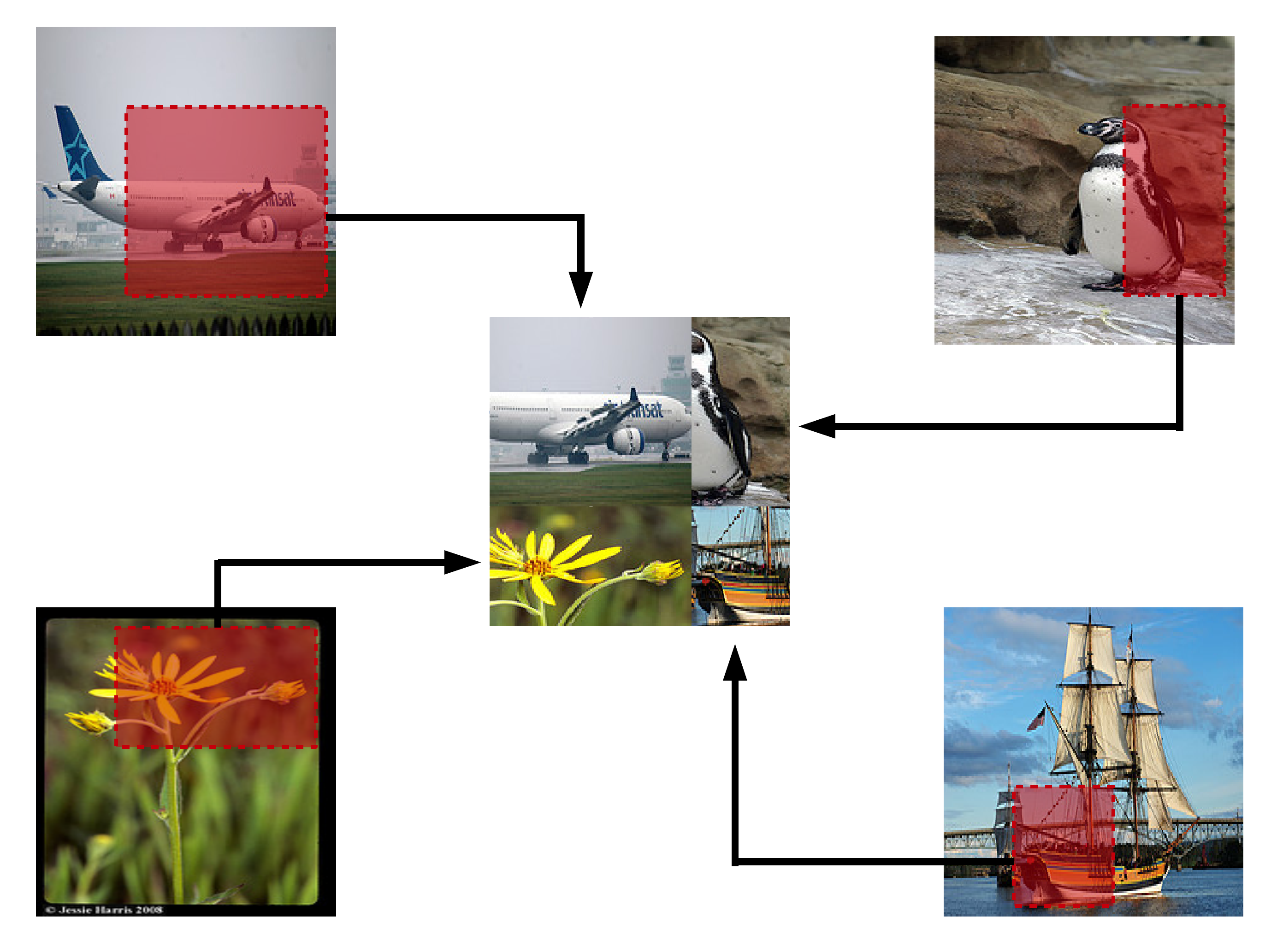}
\caption{
Conceptual explanation of the proposed \emph{random image cropping and patching} (\emph{RICAP}) data augmentation.
Four training images are randomly cropped as denoted by the red shaded areas, and patched to construct a new training image (at center).
The size of the final image is identical to that of the original one (e.g., $32\times32$ for the CIFAR dataset~\cite{Krizhevsky2009}).
These images are collected from the training set of the ImageNet dataset~\cite{Russakovsky2014}.
}
\label{fig:fusion_concept}
\end{figure}

In this study, as a further advancement in data augmentation, we propose a novel method called \emph{random image cropping and patching} (\emph{RICAP}).
RICAP crops four training images and patches them to construct a new training image; it selects images and determines the cropping sizes randomly, where the size of the final image is identical to that of the original image.
A conceptual explanation is shown in Fig.~\ref{fig:fusion_concept}.
RICAP also mixes class labels of the four images with ratios proportional to the areas of the four images.
Compared to mixup, RICAP has three clear distinctions: it mixes images spatially, it uses partial images by cropping, and it does not create features that are absent in the original dataset except for boundary patching.

We introduce the detailed algorithm of RICAP in Section~\ref{sec:proposed_method_general} and explain its conceptual contributions in Section~\ref{sec:concept}.
We apply RICAP to existing deep CNNs and evaluate them on the classification tasks using the CIFAR-10, CIFAR-100~\cite{Krizhevsky2009}, and ImageNet~\cite{Russakovsky2014} datasets in Sections~\ref{sec:c10_c100}, \ref{sec:imagenet} and \ref{sec:others}.
The experimental results demonstrate that RICAP outperforms existing data augmentation techniques,
In particular, RICAP achieves a new state-of-the-art performance on CIFAR-10 classification task.
Furthermore, in Section~\ref{sec:visualization}, we visualize the region where the model focuses attention using class activation mapping~\cite{Zhou2016}, demonstrating that RICAP makes CNNs focus attention on various objects and features in an image, in other words, RICAP prevents CNNs from overfitting to specific features.
We demonstrated in Section~\ref{sec:visualization_background} that the model can learn the deeper relationship between the object and its background when a cropped region contains no object.
We describe the ablation study performed in Section~\ref{sec:ablation} and make further comparison of RICAP with mixup in Section~\ref{sec:4mixup}.
In addition, we confirm that RICAP works well for an image-caption retrieval task using Microsoft COCO dataset~\cite{Lin2014} in Section~\ref{sec:image-caption}, person re-identification task in Section~\ref{sec:person_reid} and object detection task in Section~\ref{sec:detection}.

Limited preliminary results can be found in a recent conference proceeding~\cite{Takahashi2018ACML}.
The improvements from the proceeding are as follows.
We modified the distribution generating boundary position from uniform distribution having two parameters to beta distribution having one parameter.
This modification simplified RICAP and improved the performance.
By visualizing regions to which a CNN pays much attention, we demonstrated that RICAP supports the CNN to use wider variety of features from the same image and prevents the overfitting to features of a specific region in Section~\ref{sec:visualization}.
We also visualized that the CNN trained with RICAP learns the deeper relationship between the foreground object and its background, which is thanks to the chance of training with cropped regions containing no objects in Section~\ref{sec:visualization_background}.
We performed the ablation study to evaluate contributions of image mixing (cropping and patching) and label mixing of RICAP in Section~\ref{sec:ablation}.
We demonstrated the importance of image patching in RICAP by a detailed comparison with mixup in Section~\ref{sec:4mixup}.
We confirmed that RICAP functions well for image-caption retrieval task, person re-identification task, and object detection task in addition to the image classification in Sections~\ref{sec:image-caption},~\ref{sec:person_reid}, and~\ref{sec:detection}.
We appended a Python code of RICAP for reproduction in Algorithm~\ref{ricap_algorithm1}.


\section{Related Works}
\label{sec:Related_works}

RICAP is a novel data augmentation technique and can be applied to deep CNNs in the same manner as conventional techniques.
In addition, RICAP is related to the soft labels.
In this section, we explain related works on data augmentation and soft labels.

\subsection{Data Augmentation}
\label{sec:related_works_dataaug}

Data augmentation increases the variety of training samples and prevents overfitting~\cite{Ciresan2011,Ciresan2012}.
We introduce related methods by categorizing them into four groups; standard method, data disrupting method, data mixing method, and auto-adjustment method.

\indent
\subsubsection{Standard Data Augmentation Method}
\label{sec:standard_data_aug}

A deep CNN, AlexNet~\cite{Krizhevsky2012}, used random cropping and horizontal flipping for evaluation on the CIFAR dataset~\cite{Krizhevsky2009}.
Random cropping prevents a CNN from overfitting to specific features by changing the apparent features in an image.
Horizontal flipping doubles the variation in an image with specific orientations, such as a side-view of an airplane.
AlexNet also performed principal component analysis (PCA) on a set of RGB values to alter the intensities of the RGB channels for evaluation on the ImageNet dataset~\cite{Russakovsky2014}.
They added multiples of the found principal components to each image.
This type of color translation is useful for colorful objects, such as flowers.
Facebook AI Research employed another method of color translation called color jitter for the reimplementation of ResNet~\cite{He2016} available at \url{https://github.com/facebook/fb.resnet.torch}.
Color jitter randomly changes the brightness, contrast, and saturation of an image instead of the RGB channels.
These traditional data augmentation techniques play an important role in training deep CNNs.
However, the number of parameters is ever-growing and the risk of overfitting is also ever-increasing as many studies propose new network architectures~\cite{He2016b,Zagoruyko2016,Huang2016b,Han2016,Xie2017} following ResNet\cite{He2016}.
Therefore, data augmentation techniques have attracted further attention.

\indent
\subsubsection{Data Disrupting Method}
\label{sec:disturb_data_aug}
The aforementioned standard methods simply enrich datasets because the resultant images are still natural.
Contrary to them, some methods produce unnatural images by disrupting images' features.
Dropout on the input layer~\cite{Hinton2012} is a data augmentation technique~\cite{Zhao2019a} that disturbs and masks the original information of given data by dropping pixels.
Pixel dropping functions as injection of noise into an image~\cite{Sietsma1991}.
It makes the CNN robust to noisy images and contributes to generalization rather than enriching the dataset.

Cutout randomly masks a square region in an image at every training step~\cite{DeVries2017}.
It is an extension of dropout, where masking of regions behaves like injected noise and makes CNNs robust to noisy images.
In addition, cutout can mask the entire main part of an object in an image, such as the face of a cat.
In this case, CNNs need to learn other parts that are usually ignored, such as the tail of the cat.
This prevents deep CNNs from overfitting to features of the main part of an object.
A similar method, random erasing, has been proposed~\cite{Zhong2017a}.
It also masks a certain area of an image but has clear differences; it randomly determines whether to mask a region as well as the size and aspect ratio of the masked region.

\indent
\subsubsection{Data Mixing Method}
\label{sec:mixing_data_aug}
This is a special case of data disrupting methods.
Mixup alpha-blends two images to construct a new training image~\cite{Zhang2017}.
Mixup can train deep CNNs on convex combinations of pairs of training samples and their labels, and enables deep CNNs to favor a simple linear behavior in-between training samples.
This behavior makes the prediction confidence transit linearly from one class to another class, thus providing smoother estimation and margin maximization.
Alpha-blending not only increases the variety of training images but also works like adversarial perturbation~\cite{Goodfellow2015a}.
Thereby, mixup makes deep CNNs robust to adversarial examples and stabilizes the training of generative adversarial networks.
In addition, it behaves as soft labels by mixing class labels with the ratio $\lambda:1-\lambda$~\cite{Szegedy2016}.
We explain soft labels in detail below.

\indent
\subsubsection{Auto-adjustment Method}
\label{sec:method_data_aug}
AutoAugment~\cite{Cubuk2018} is a framework exploring the best hyperparameters of existing data augmentations using reinforcement learning~\cite{Zoph2016}.
Hence, it is not a data augmentation method but is an external framework.
It achieved significant results on the CIFAR-10 classification and proved the importance of data augmentation for the learning of deep CNN.

\subsection{Soft Labels}
\label{sec:soft_label}
For classification tasks, the ground truth is typically given as probabilities of $0$ and $1$, called the hard labels~\cite{Hinton2014a}.
A CNN commonly employs the softmax function, which never predicts an exact probability of $0$ and $1$.
Thus, the CNN continues to learn increasingly larger weight parameters and make an unjustly high confidence.
Knowledge distillation~\cite{Hinton2014a} proposed to use soft labels, which have intermediate probabilities such as $0.1$ and $0.9$.
This method employs the predictions of a trained CNN using the softmax function with a high temperature to obtain the soft labels.
The soft labels contain rich information about the similarity between classes and the ambiguity of each sample (For example, the dog class is similar to the cat class rather than the plane class).
As a simpler approach, Szegedy et al.~proposed label smoothing, which provides the soft labels of given probabilities such as $0.9$ and $0.8$~\cite{Szegedy2016}.
The label smoothing prevents the endless pursuit of hard $0$ and $1$ probabilities for the estimated classes and enables the weight parameters to converge to certain values without discouraging correct classification.
Mixup provides the soft labels of a probability equal to the $\alpha$-value of the blending, which regularizes the CNN to favor a simple linear behavior in-between training images because mixup blends two images at the pixel level.
As a result, mixup pushes the decision boundary away from original samples and maximizes the margin.


\section{Proposed Method}
\label{sec:proposed_method}

\subsection{Random Image Cropping and Patching (RICAP)}
\label{sec:proposed_method_general}

In this study, we propose a novel data augmentation technique called \emph{random image cropping and patching} (\emph{RICAP}) for deep convolutional neural networks (CNNs).
The conceptual explanation of RICAP is shown in Fig.~\ref{fig:fusion_concept}.
It consists of three data manipulation steps.
First, four images are randomly selected from the training set.
Second, the images are cropped separately.
Third, the cropped images are patched to create a new image.
Despite this simple procedure, RICAP increases the variety of images drastically and prevents overfitting of deep CNNs having numerous parameters.

\begin{figure}[!t]{}
\centering
\includegraphics[width=3.5in,bb= 0 0 740 580,clip]{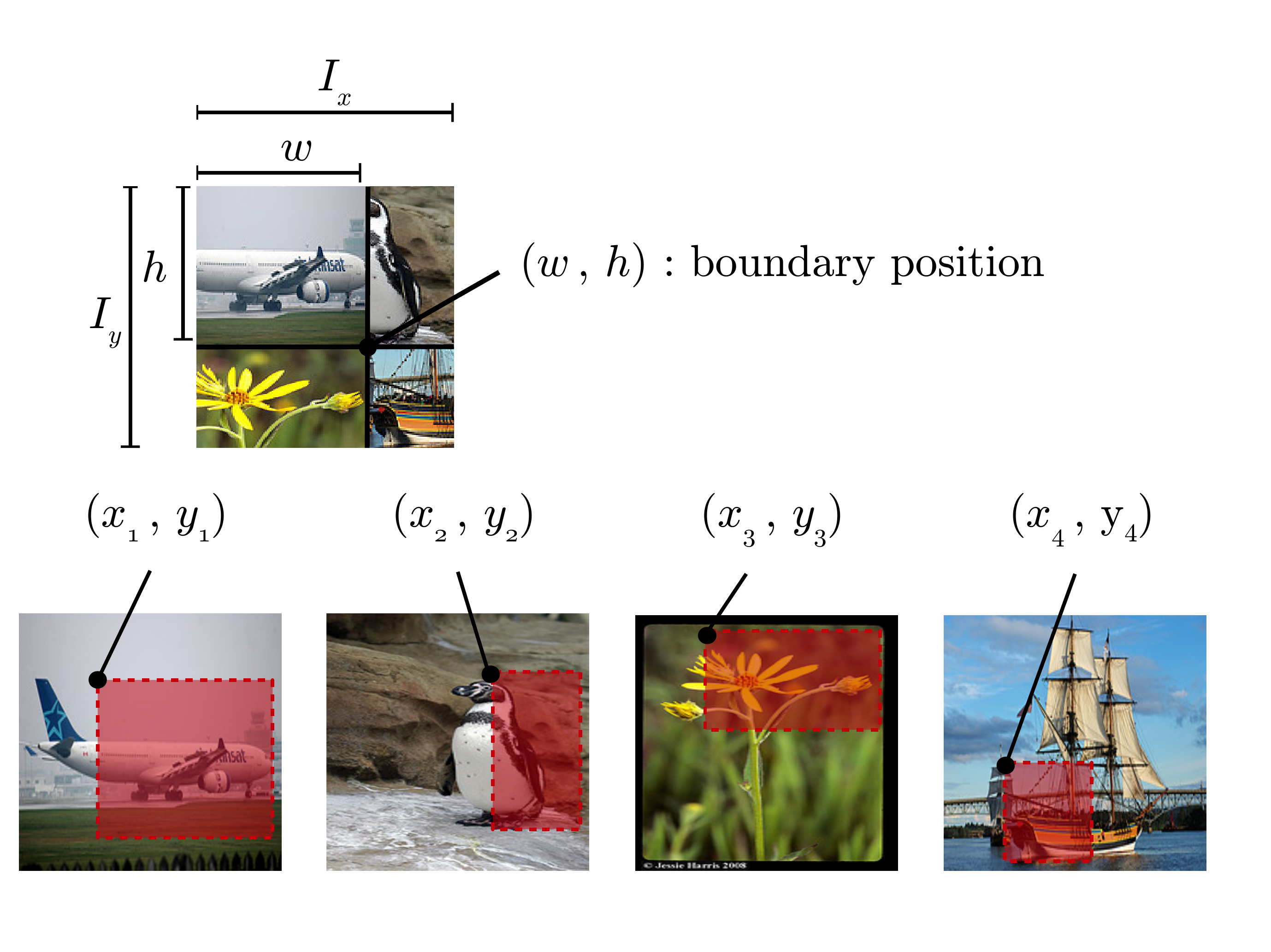}
\caption{
Detailed explanation of RICAP.
$I_x$ and $I_y$ are the width and height of the original image, respectively.
Four images are randomly cropped, as denoted by the red shaded areas, and patched according to the boundary position $(w,h)$.
The boundary position $(w,h)$ is generated from a beta distribution $\textrm{Beta}(\beta,\beta)$, where $\beta$ is a hyperparameter of RICAP.
Based on the boundary position $(w,h)$, the cropped positions $(x_k,y_k)$ are selected such that they do not change the image size.
}
\label{fig:fusion_implementation}
\end{figure}

A more specific explanation of the implementation is shown in Fig.~\ref{fig:fusion_implementation}.
We randomly select four images $k\in\{1,2,3,4\}$ from the training set and patch them on the upper left, upper right, lower left, and lower right regions.
$I_x$ and $I_y$ denote the width and height of the original training image, respectively.
$(w,h)$ is the boundary position which gives the size and position of each cropped image.
We choose this boundary position $(w,h)$ in every training step from beta distributions as below.
\begin{equation*}
  \begin{array}{ll}
  w= \mathrm{round}(w'I_x),&
  h= \mathrm{round}(h'I_y),\\
  w'\sim \textrm{Beta}(\beta,\beta),&
  h'\sim \textrm{Beta}(\beta,\beta),
  \end{array}
\end{equation*}
where $\beta \in (0,\infty)$ is a hyperparameter and $\mathrm{round}(\cdot)$ is the rounding function.
Once we determine the boundary position $(w,h)$, we automatically obtain the cropping sizes $(w_k, h_k)$ of the images $k$, i.e., $w_1=w_3=w$, $w_2=w_4=I_x-w$, $h_1=h_2=h$, and $h_3=h_4=I_y-h$.
For cropping the four images $k$ following the sizes $(w_k,h_k)$, we randomly determine the positions $(x_k,y_k)$ of the upper left corners of the cropped areas as
\begin{align*}
  x_k&\sim \mathcal{U}(0, I_x - w_k),\\
  y_k&\sim \mathcal{U}(0, I_y - h_k).
\end{align*}

\subsection{Label Mixing of RICAP for Classification}
\label{sec:label_mix}
For the classification task, the class labels of the four images are mixed with ratios proportional to the image areas.
We define the target label $c$ by mixing one-hot coded class labels $c_k$ of the four patched images with ratios $W_i$ proportional to their areas in the new constructed image;
\begin{equation}
  c = \sum_{k\in\{1,2,3,4\}} W_k c_k\;\;\mathrm{for}\;\;W_k = \frac{w_k h_k}{I_x I_y},\label{eq:mixweight}
\end{equation}
where $w_k h_k$ is the area of the cropped image $k$ and $I_x I_y$ is the area of the original image.

\subsection{Hyperparameter of RICAP}
\label{sec:hyperparameter}
The hyperparameter $\beta$ determines the distribution of boundary position.
If $\beta$ is large, the boundary position $(w,h)$ tends to be close to the center of a patched image and the target class probabilities $c$ often have values close to $0.25$.
RICAP encounters a risk of excessive soft labeling, discouraging correct classification.
If $\beta$ is small, the boundary position $(w,h)$ tends to be close to the four corners of the patched image and the target class probabilities $c$ often have $0$ or $1$ probabilities.
Especially, with $\beta=0$, RICAP does not augment images but provides original images.
With $\beta=1.0$, the boundary position $(w,h)$ is distributed uniformly over the patched images.
For reproduction, we provide a Python code of RICAP for classification in Algorithm~\ref{ricap_algorithm1} in Appendix.

\subsection{Concept of RICAP}
\label{sec:concept}
RICAP shares concepts with cutout, mixup, and soft labels, and potentially overcomes their shortcomings.

Cutout masks a subregion of an image and RICAP crops a subregion of an image.
Both change the apparent features of the image at every training step.
However, masking in cutout simply reduces the amount of available features in each sample.
Conversely, the proposed RICAP patches images, and hence the whole region of a patched image produces features contributing to the training.

Mixup employs an alpha-blend (i.e., blending of pixel intensity), while RICAP patches four cropped images, which can be regarded as a spatial blend.
By alpha-blending two images, mixup generates pixel-level features that original images never produce, drastically increasing the variety of features that a CNN has to learn and potentially disrupting the training.
Conversely, images patched by RICAP method always produce pixel-level features that original images also produce except for boundary patching.

\begin{figure}[!t]{}
\centering
\includegraphics[width=3.8in,bb= 0 0 840 600,clip]{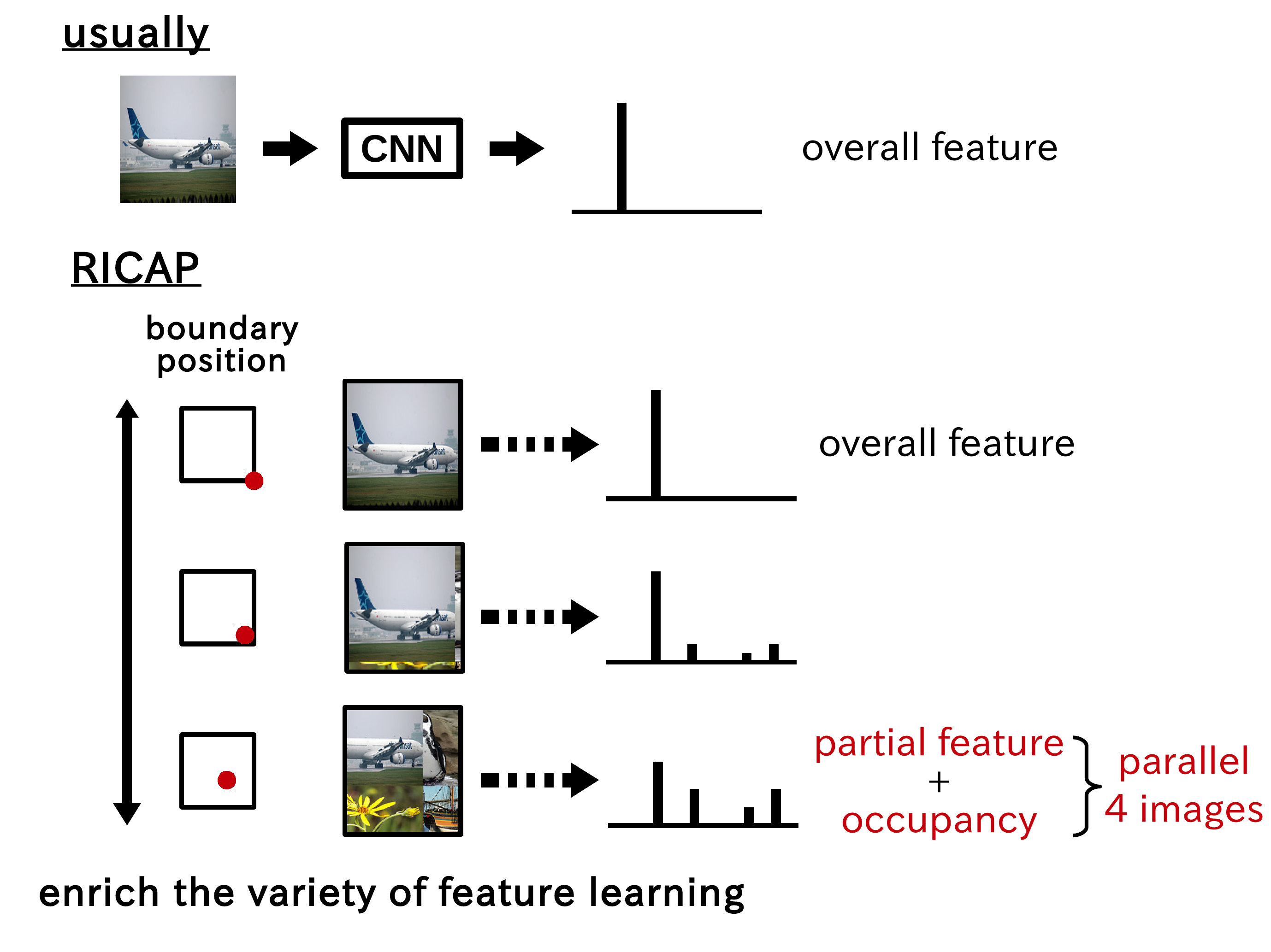}
\caption{
Comparison between classification by usual CNN training and by RICAP training.
Based on the boundary position, image mixing and soft labels of RICAP changes its role.
In the case of the boundary position is close to four corners, CNN learns the overall features or enjoy the benefit of the soft labels.
In the case of the boundary position close to center of patched image, RICAP replaces the classification task with the occupancy estimation task.
This occupancy estimation forces the CNN to classify each pixel, and thereby, the CNN becomes to use minor features, partial features, backgrounds, and any other information that is often ignored in parallel $4$ images.
}
\label{fig:ricap_work}
\end{figure}

The label smoothing always provides soft labels for preventing endless pursuit of the hard probabilities.
Also, mixup provides the soft labels of a probability equal to the $\alpha$-value of the blending, leading the decision boundary away from original samples and the margin maximization.
On the other hand, RICAP replaces the classification task with the occupancy estimation task by mixing the four class labels with ratios proportional to the areas of the four cropped images.
This indicates that RICAP forces the CNN to classify each pixel in a weakly supervised manner, and thereby, the CNN becomes to use minor features, partial features, backgrounds, and any other information that is often ignored.
In particular, the extreme case that the cropped area has no object is described in Section~\ref{sec:no_object}.
RICAP tends to provide softer labels than the label smoothing, so that the soft labels of RICAP without the image patching disturbs the classification as shown in Section~\ref{sec:ablation}.
These studies share the sense of soft labels but their contributions are vastly different.
We summarized the works of RICAP by image mixing and label mixing on image classification in the Fig.~\ref{fig:ricap_work}.

\subsection{Object Existence in Cropped Areas}
\label{sec:no_object}
When the boundary position $(w,h)$ is close to the four corners, a cropped area becomes small and occasionally depicts no object.
For classifying natural images, we expect that a part of the subject is basically cropped, but of course, the cropped region could contain no object by cropping only the background.
In this case, a CNN tries to associate the background with the subject class because the CNN has to output the posterior probability of the subject class proportional to the area of the cropped region.
Thereby, the CNN learns the relationship between the object class and the background.
Hence, the chance of RICAP that the cropped region contains no object does not limit the performance of the CNN but improves it.
When two classes share similar backgrounds (e.g., planes and birds are depicted in front of the sky), the backgrounds are associated with both classes as distinguished from other classes.
Moreover, when a cropped region is too small to learn features, the CNN simply ignores the cropped region and enjoys the benefit of the soft labels like the label smoothing.
We will evaluate this concept in Section~\ref{sec:visualization_background}.

\subsection{Differences between RICAP and mixup}
\label{sec:dif_ricap_mixup}
\begin{figure}[!t]{}
\centering
\includegraphics[width=2.5in,bb= 150 0 670 580,clip]{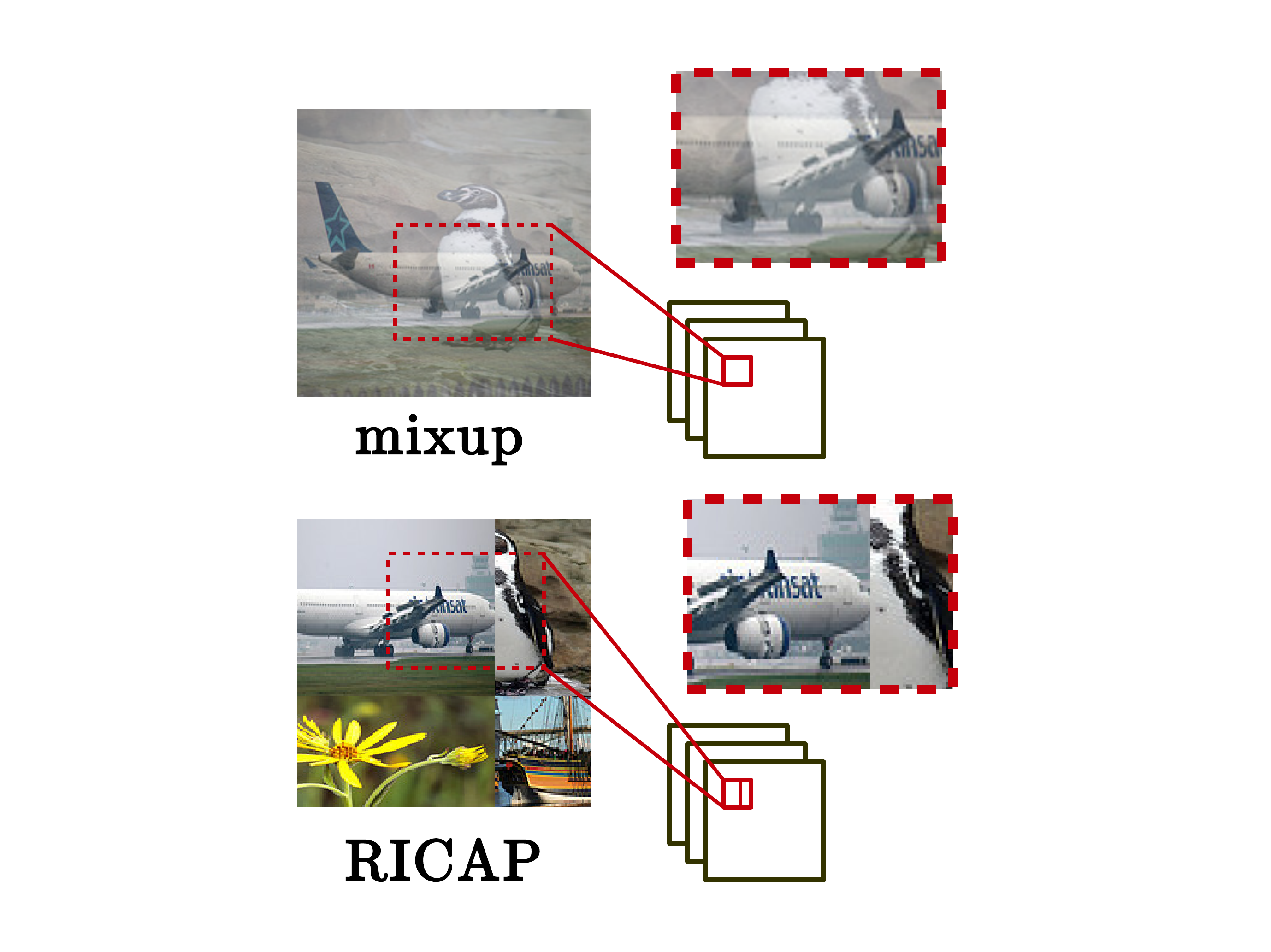}
\caption{
Comparison between images processed by RICAP and mixup.
}
\label{fig:ricap_and_mixup}
\end{figure}

Before we end Section~\ref{sec:proposed_method}, we summarize again the main differences between RICAP and mixup to emphasize the novelty of RICAP by using Fig.~\ref{fig:ricap_and_mixup}.

A main difference is the blending strategy; RICAP employs patching (i.e., spatial blending) while mixup employs alpha-blending (i.e., pixel-wise blending), as shown in the left two images in Fig.~\ref{fig:ricap_and_mixup}.
To clarify this impact, we focus on the sub-areas surrounded by the red dotted lines.
As depicted in the upper right panel, by blending two objects, mixup's alpha-blending sometimes creates local features that are absent in the original dataset and leads to an extremely difficult recognition task.
This tendency disturbs the model training as excessive adversarial perturbation, or at least, wastes the computational time and model capacity.
On the other hand, as depicted in the lower right panel, RICAP's patching (spatial blending) always produces local features included in the original dataset.
The local features always support the model training.
Of course, the patching can create new global features by combining multiple objects, but this tendency prevents the CNN from overfitting to the existing combination of objects.

Another main difference is the cropping by RICAP.
As shown in the upper left panel, the whole bodies of a penguin and an aircraft are still recognizable even after mixup's alpha-blending because the backgrounds are simple textures.
Alpha-blending an object with a background is insufficient for masking an object, and a CNN can focus on and overfit to salient features such as penguin head or aircraft empennage.
On the other hand, RICAP's cropping removes many parts of an object, that is, removes many features literally.
Thereby, RICAP prevents the CNN from overfitting to salient features like data disrupting methods introduced in Section~\ref{sec:disturb_data_aug}.
Even when a foreground object is totally removed, the CNN is trained to find the relationship between the class labels and backgrounds.

In short, mixup tends to produce too easy or too difficult tasks while RICAP works as an appropriate regularizer.

\indent
\section{Experiments on Image Classification}
\label{sec:exp}

To evaluate the performance of RICAP, we apply it to deep CNNs and evaluate it on the classification task in this section

\begin{figure*}[!t]
\centering
\includegraphics[width=3.5in,bb= 0 0 830 600,clip]{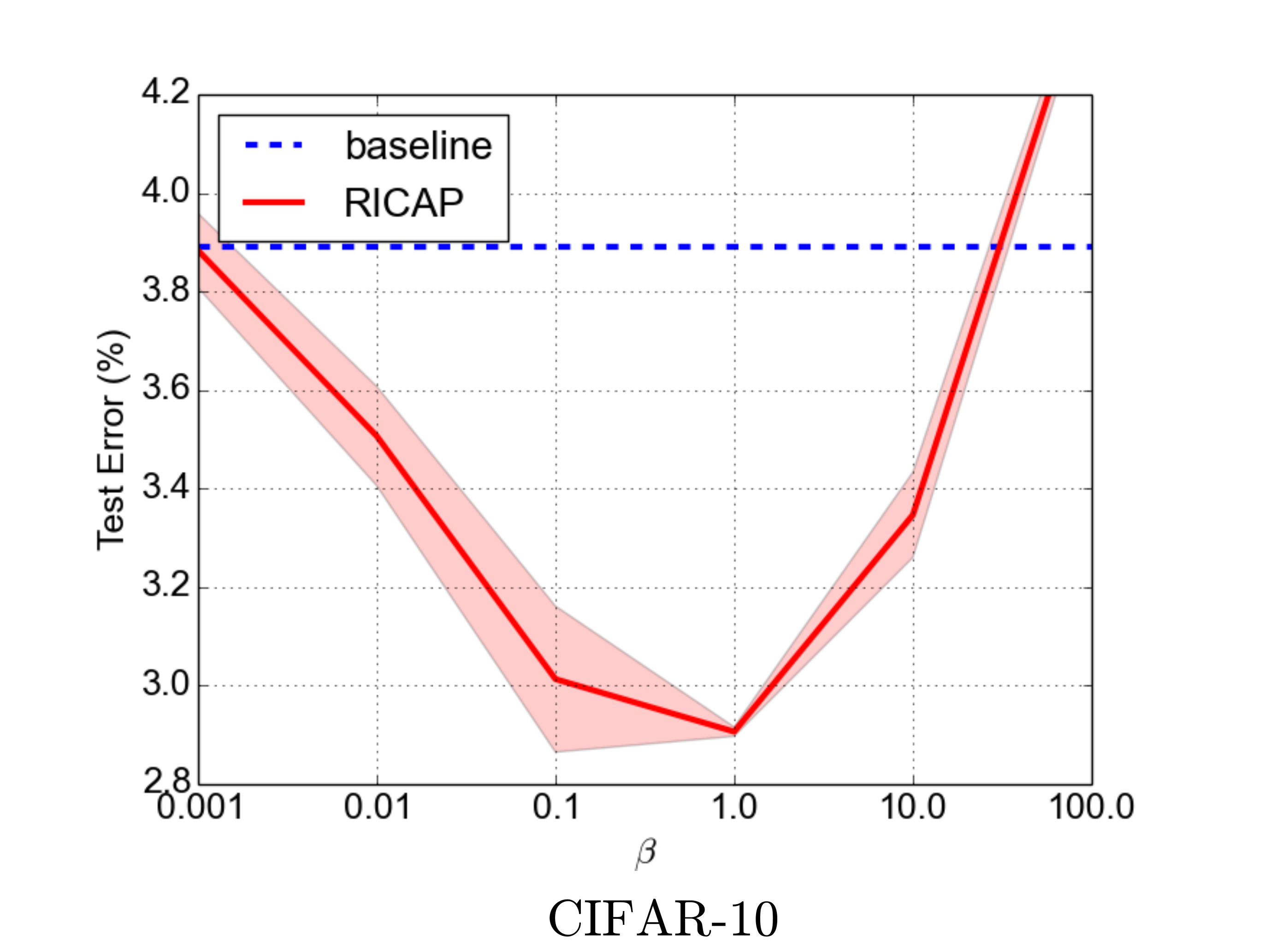}
\hspace*{-1.1cm}
\includegraphics[width=3.5in,bb= 0 0 830 600,clip]{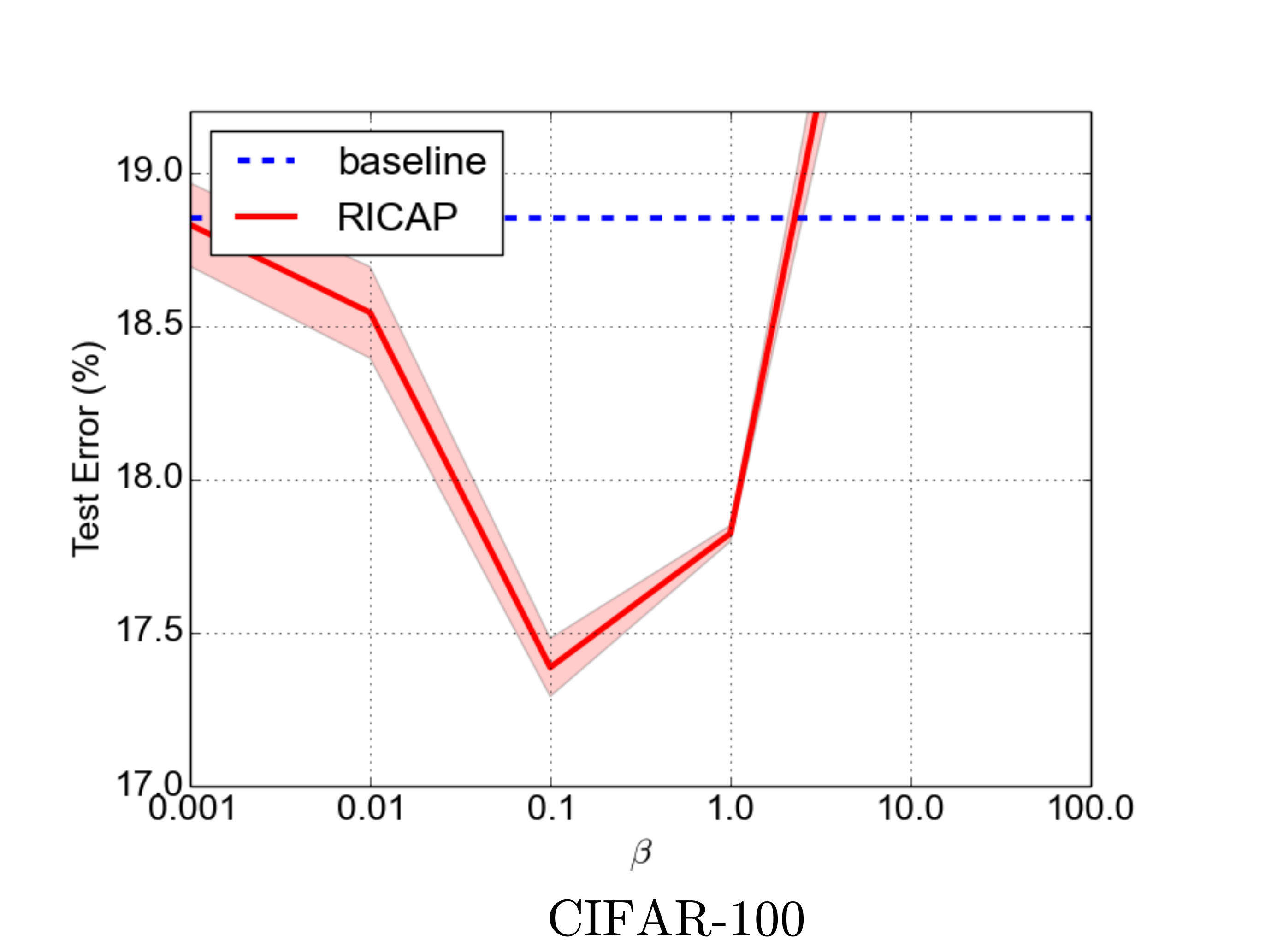}\\
\includegraphics[width=3.5in,bb= 0 0 830 600,clip]{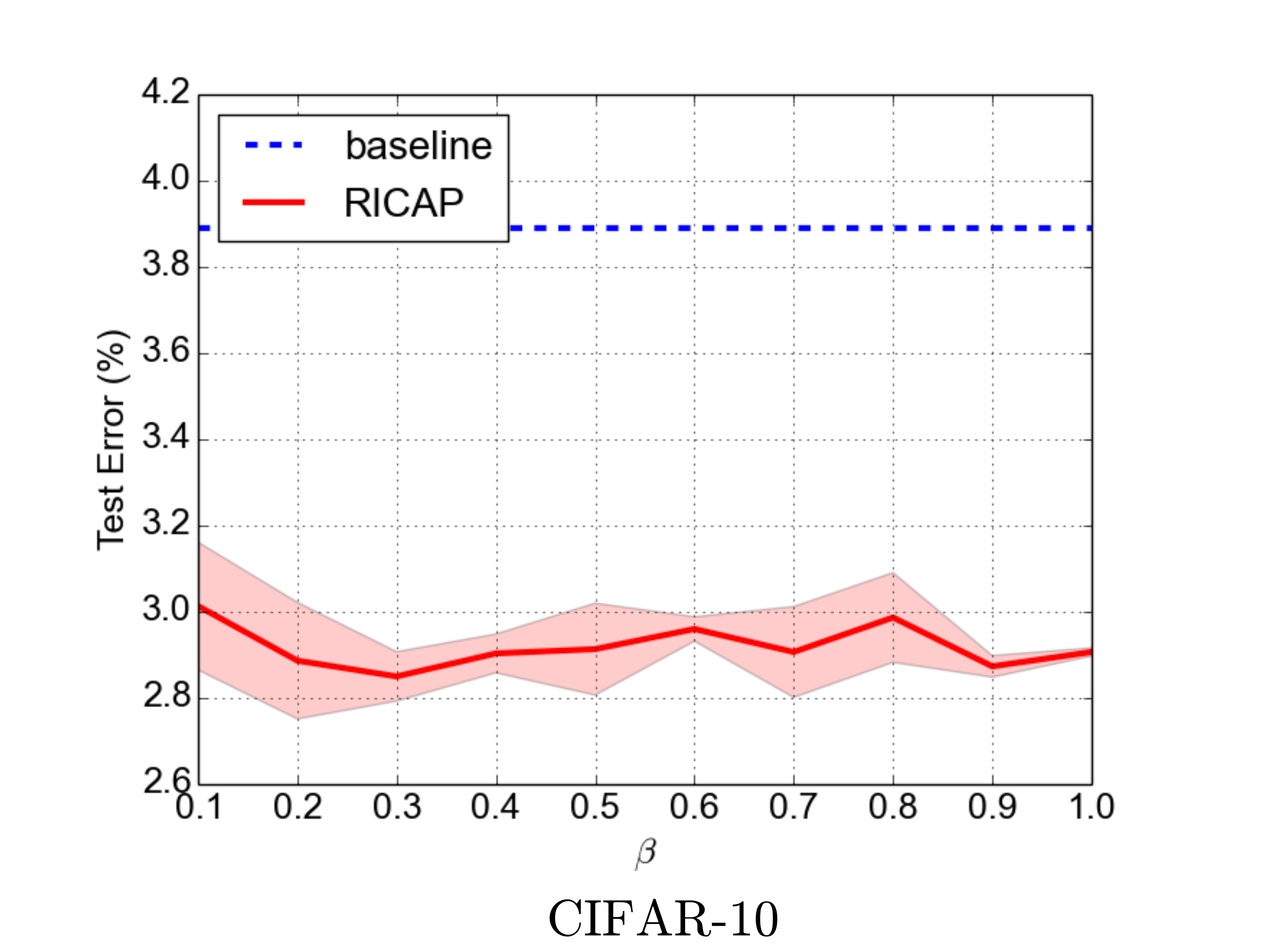}
\hspace*{-1.1cm}
\includegraphics[width=3.5in,bb= 0 0 830 600,clip]{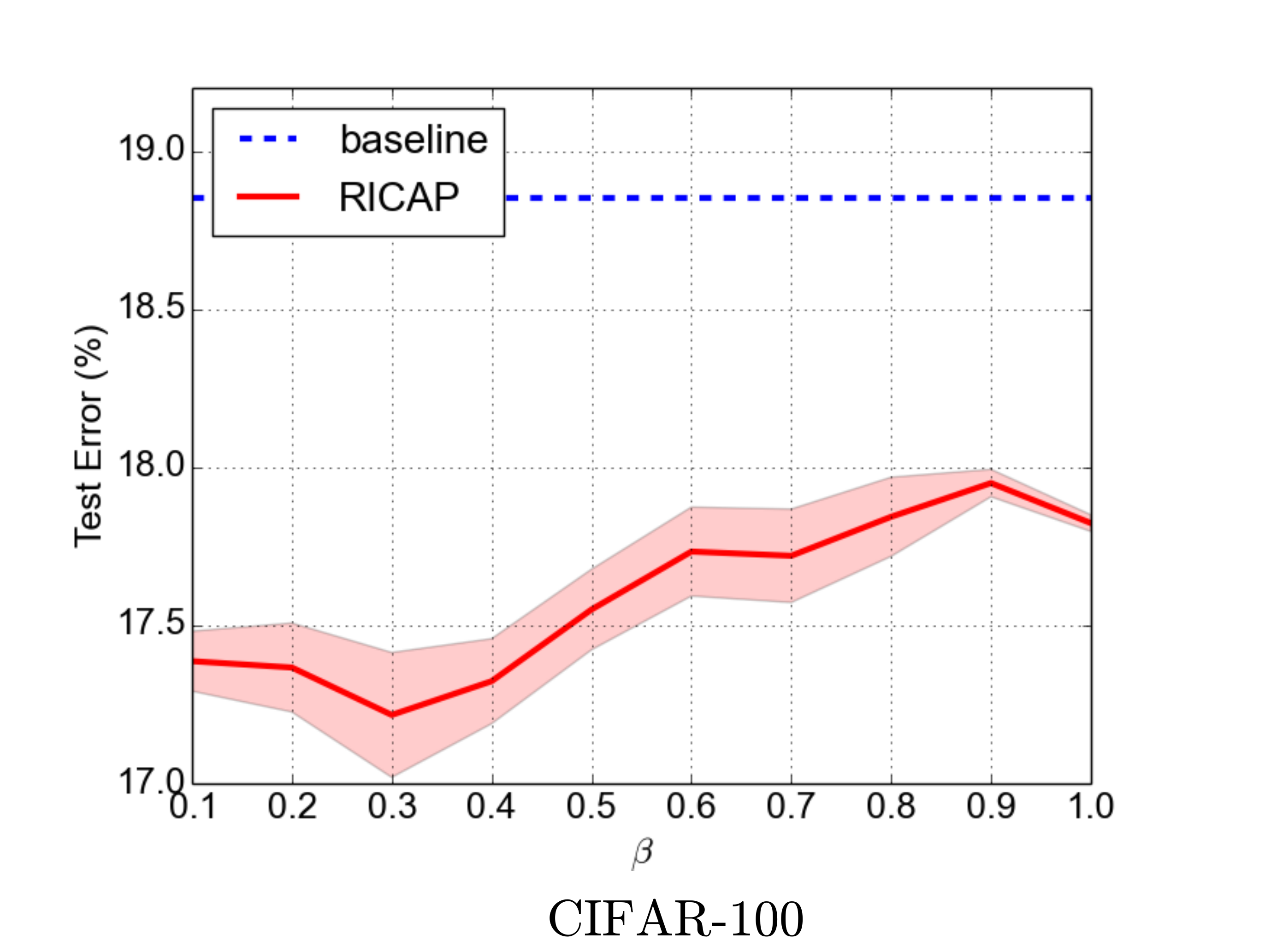}\\
\caption{
Exploration of the hyperparameter $\beta$ of RICAP using the WideResNet 28-10 for a wider range of $\beta$ on CIFAR-10 (left upper panel) and on CIFAR-100 (right upper panel) and for a more specific range of $[0.1,1.0]$ on CIFAR-10 (left lower panel) and on CIFAR-100 (right upper panel).
We performed three runs, depicting the means and standard deviations by solid lines and shaded areas, respectively.
The baseline indicates the results of the WideResNet without RICAP.
}
\label{fig:fusion-opt}
\end{figure*}

\indent
\subsection{Classification of CIFAR-10 and CIFAR-100}
\label{sec:c10_c100}

\paragraph*{\textbf{Experimental Settings}}
In this section, we show the application of RICAP to an existing deep CNN and evaluate it on the classification tasks of the CIFAR-10 and CIFAR-100 datasets~\cite{Krizhevsky2009}.
CIFAR-10 and CIFAR-100 consist of $32\times32$ RGB images of objects in natural scenes.
$50,000$ images are used for training and $10,000$ for test.
Each image is manually assigned one of the $10$ class labels in CIFAR-10 and one of the $100$ in CIFAR-100.
The number of images per class is thus reduced in CIFAR-100.
Based on previous studies~\cite{Lee2014,Romero2015,Springenberg2015}, we normalized each channel of all images to zero mean and unit variance as preprocessing.
We also employed $4$-pixel padding on each side, $32\times32$ random cropping, and random flipping in the horizontal direction as conventional data augmentation techniques.

We used a residual network called WideResNet\cite{Zagoruyko2016}.
We used an architecture called \emph{WideResNet 28-10}, which consists of $28$ convolution layers with a widen factor of $10$ and employs dropout with a drop probability of $p=0.3$ in the intermediate layers.
This architecture achieved the highest accuracy on the CIFAR datasets in the original study~\cite{Zagoruyko2016}.
The hyperparameters were set to be the same as those used in the original study.
Batch normalization~\cite{Ioffe2015} and ReLU activation function~\cite{Nair2010} were used.
The weight parameters were initialized following the He algorithm~\cite{He2016a}.
The weight parameters were updated using the momentum SGD algorithm with a momentum parameter of 0.9 and weight decay of $10^{-4}$ over 200 epochs with batches of $128$ images.
The learning rate was initialized to $0.1$, and then, it was reduced to $0.02$, $0.004$ and $0.0008$ at the $60$th, $120$th and $160$th epochs, respectively.

\paragraph*{\textbf{Classification Results}}
We evaluated RICAP with WideResNet to explore the best value of the hyperparameter $\beta$.
$I_x$ and $I_y$ were $32$ for the CIFAR datasets.
Fig.~\ref{fig:fusion-opt} shows the results on CIFAR-10 and CIFAR-100.
The baselines denote the results of the WideResNet without RICAP.
For both CIFAR-10 and CIFAR-100, $\beta=0.3$ resulted the best test error rates.
With an excessively large $\beta$, we obtained worse results than the baseline, which suggests the negative influence of excessive soft labeling.
With decreasing $\beta$, the performance converged to the baseline results.
We also summarized the results of RICAP in Table~\ref{tab:RICAP_WideResNet} as well as the results of competitive methods: input dropout~\cite{Hinton2012}, cutout~\cite{DeVries2017}, random erasing~\cite{Zhong2017a}, and mixup~\cite{Zhang2017}.
Competitive results denoted by $\dag$ symbols were obtained from our experiments and the other results were cited from the original studies.
In our experiments, each value following the $\pm$ symbol was the standard deviation over three runs.
Recall that WideResNet usually employs dropout in intermediate layers.
As the dropout data augmentation, we added dropout to the input layer for comparison.
The drop probability was set to $p=0.2$ according to the original study~\cite{Hinton2012}.
For other competitive methods, we set the hyperparameters to values with which the CNNs achieved the best results in the original studies: cutout size $16\times16$ (CIFAR-10) and $8\times8$ (CIFAR-100) for cutout and $\alpha=1.0$ for mixup.
RICAP clearly outperformed the competitive methods.

\begin{figure*}[!t]
\centering
\begin{tabular}{cc}
\includegraphics[width=2.5in,bb= 0 0 560 420,clip]{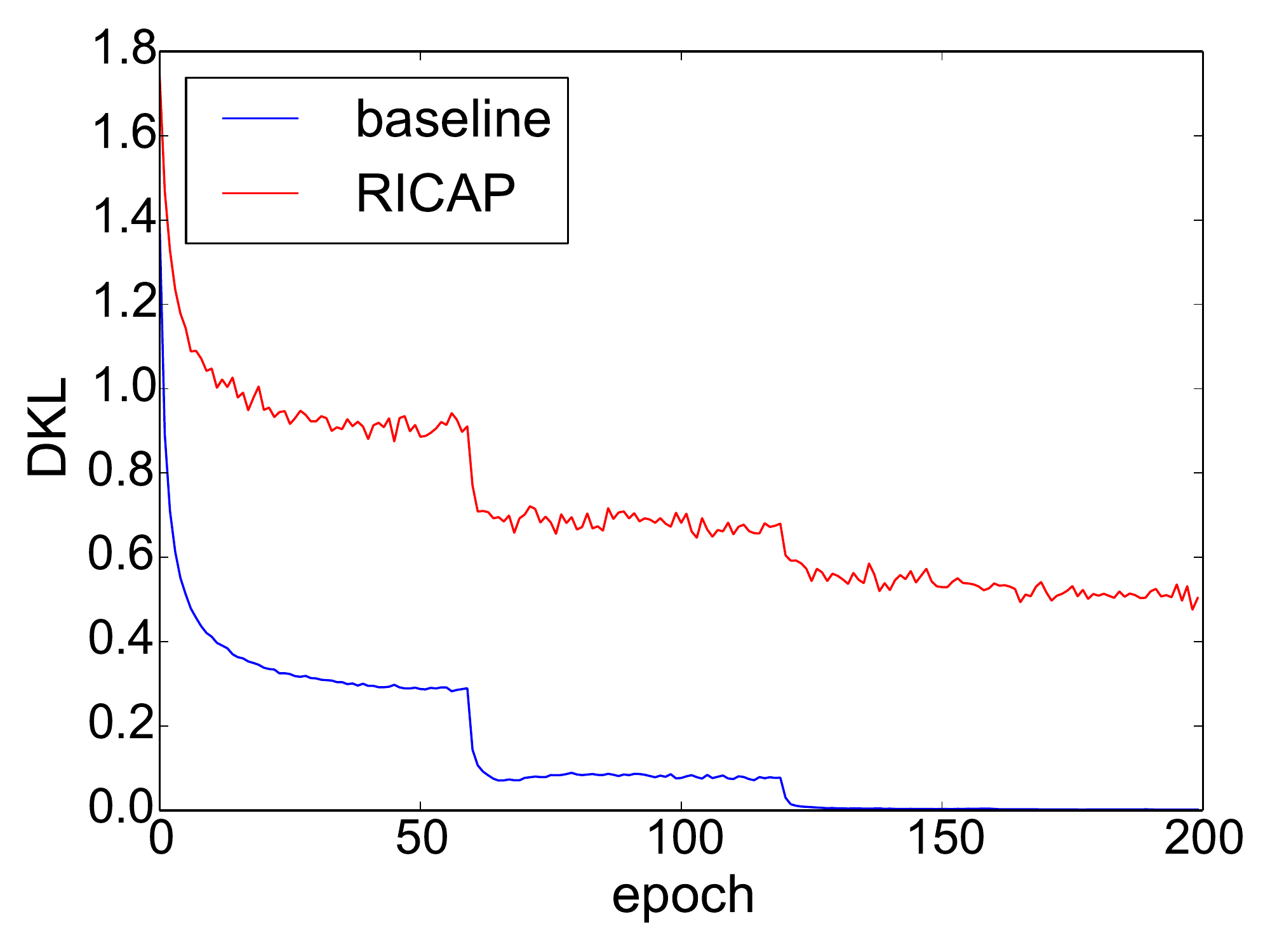}&
\includegraphics[width=2.5in,bb= 0 0 560 420,clip]{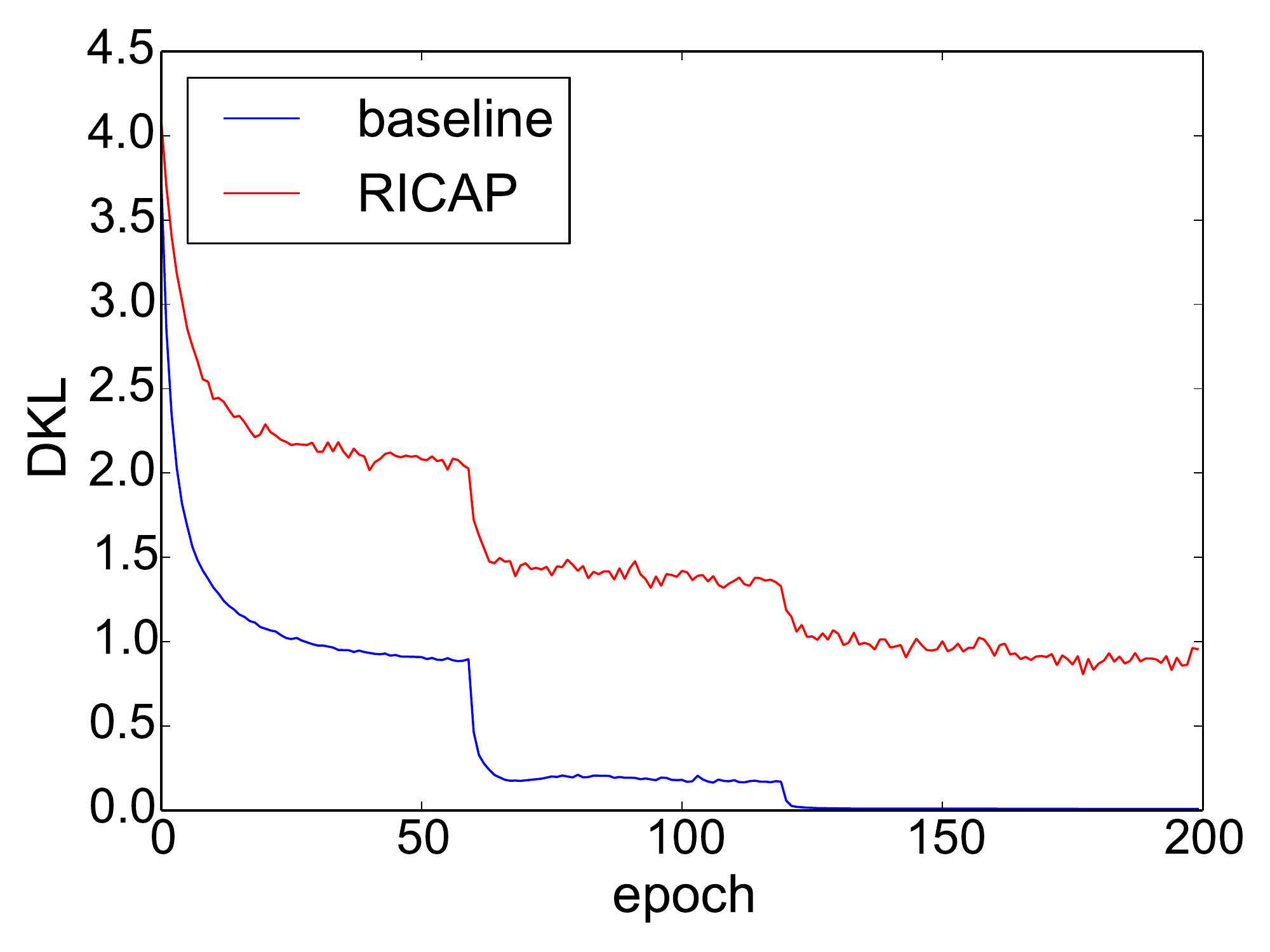}\\
\vspace*{0.4cm}
(a) Training loss on CIFAR-10 &
(b) Training loss on CIFAR-100 \\
\includegraphics[width=2.5in,bb= 0 0 560 420,clip]{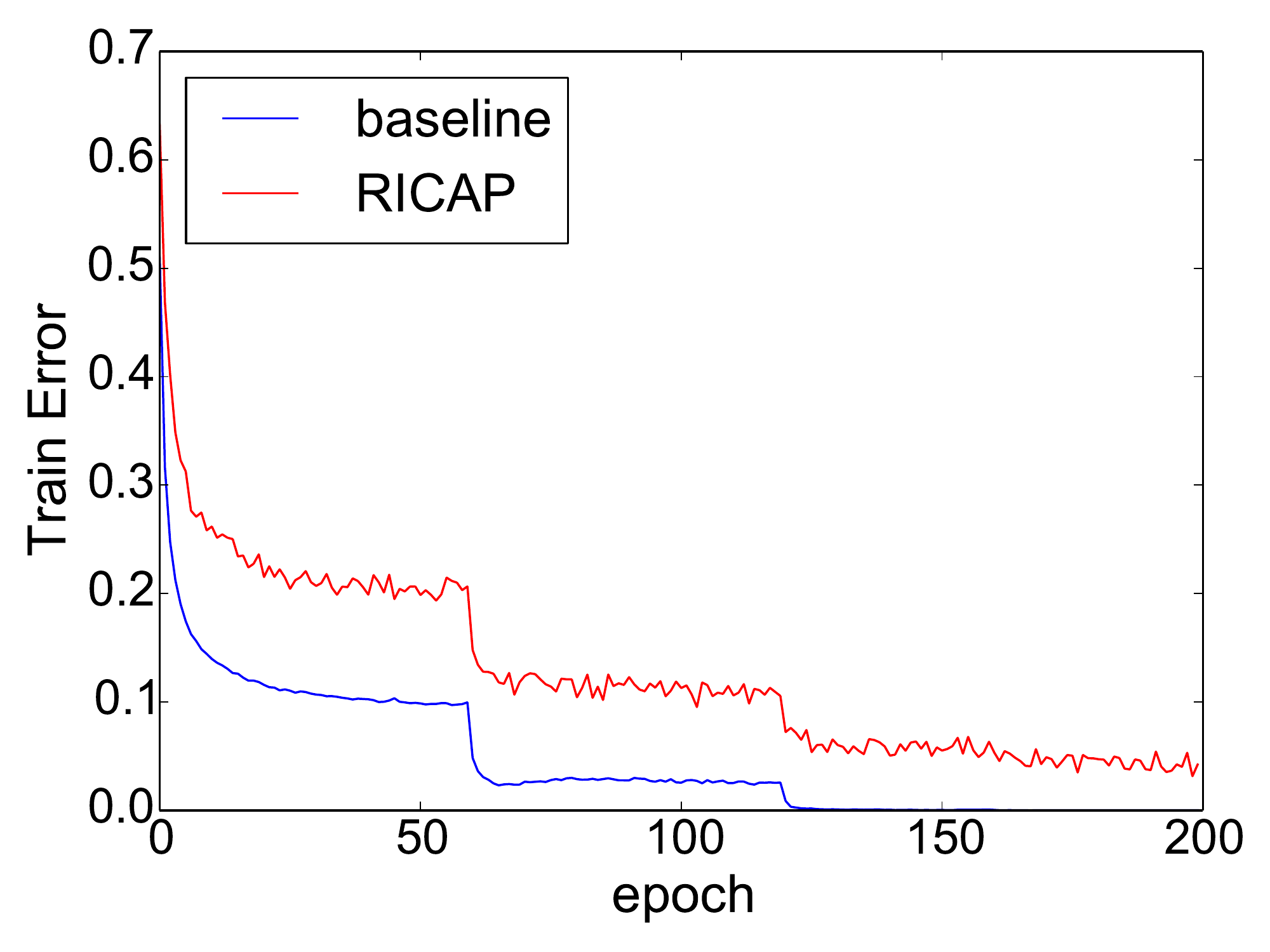}&
\includegraphics[width=2.5in,bb= 0 0 560 420,clip]{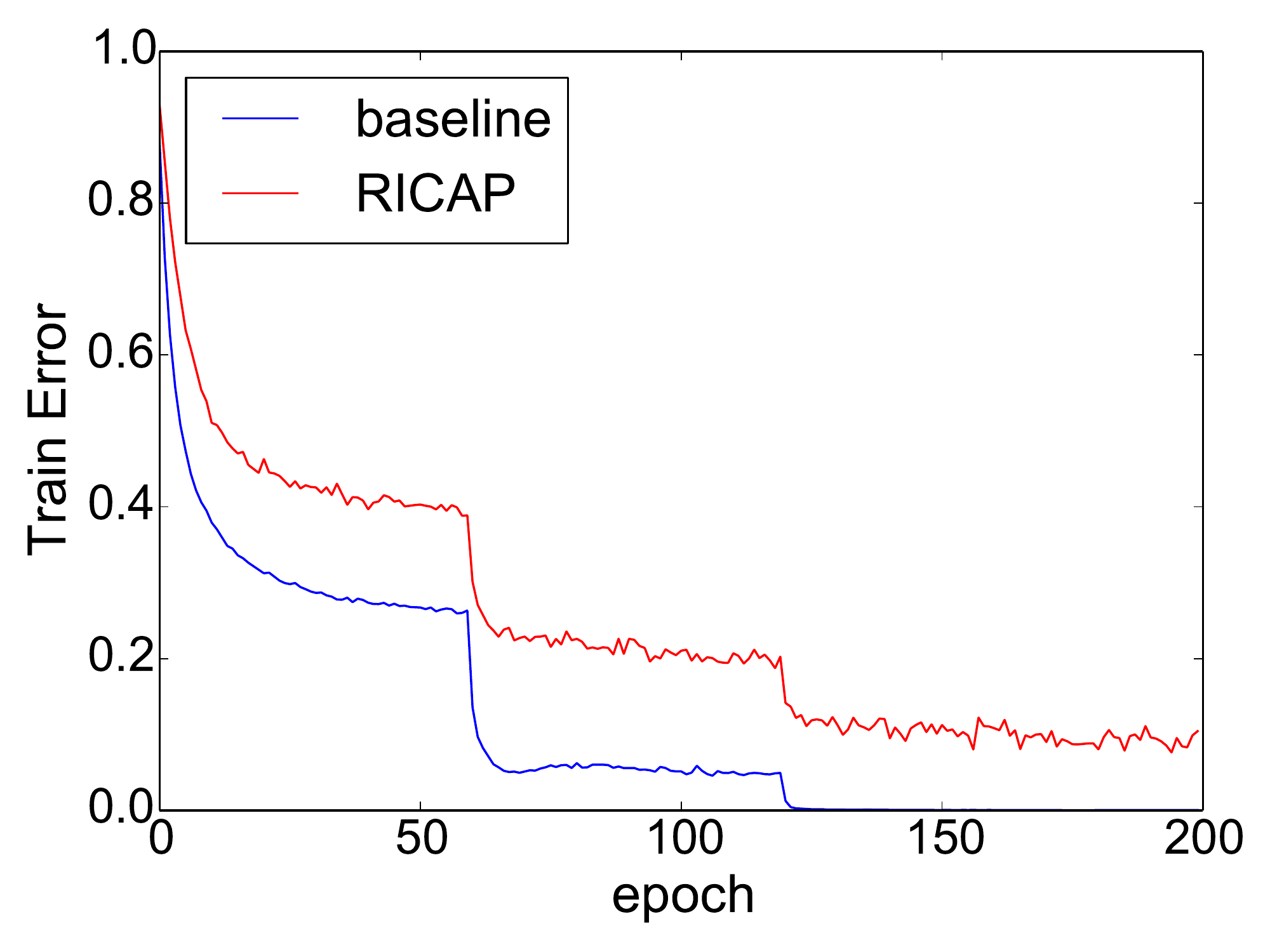}\\
\vspace*{0.4cm}
(c) Training error rate on CIFAR-10 &
(d) Training error rate on CIFAR-100 \\
\includegraphics[width=2.5in,bb= 0 0 560 420,clip]{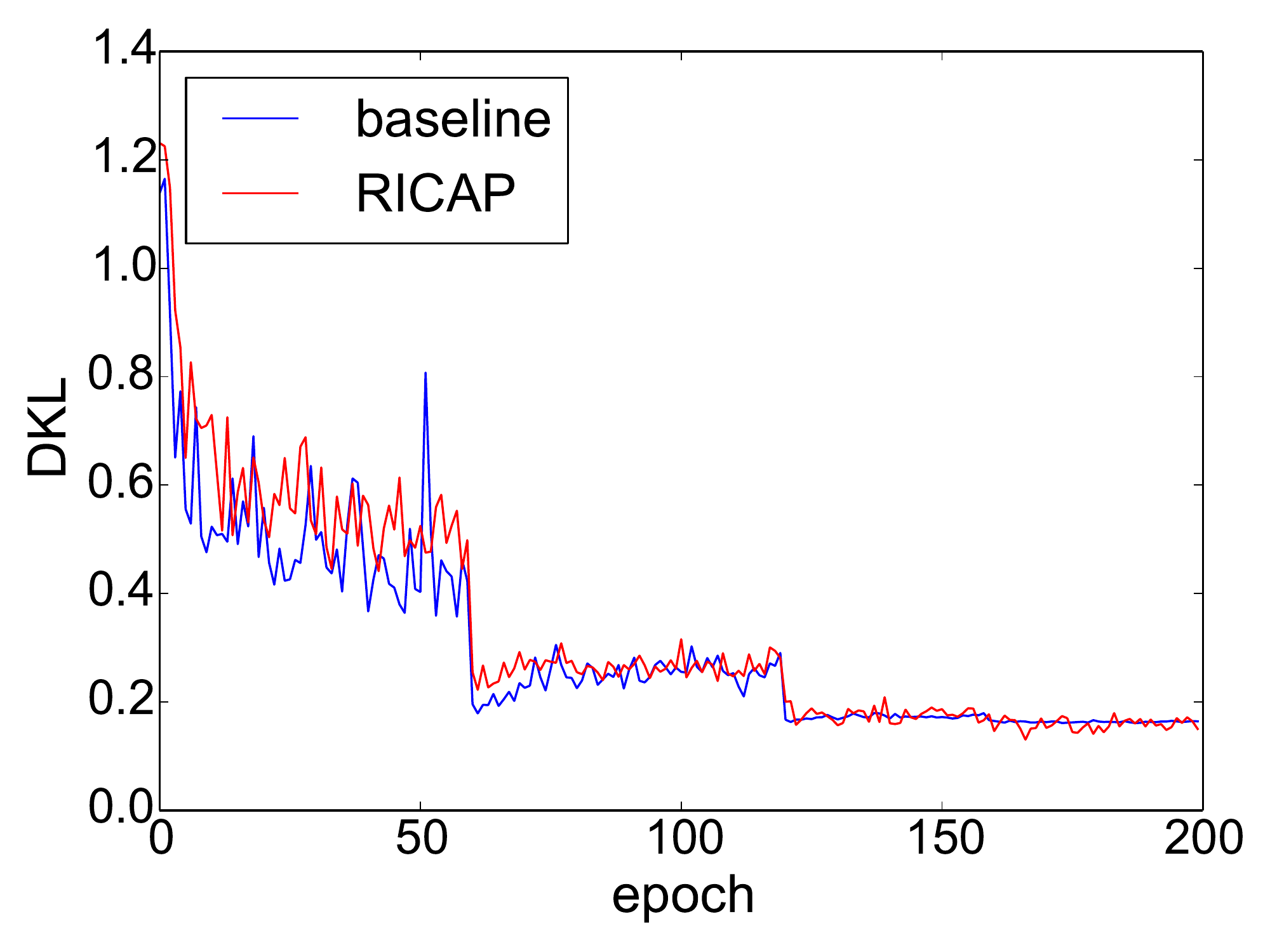}&
\includegraphics[width=2.5in,bb= 0 0 560 420,clip]{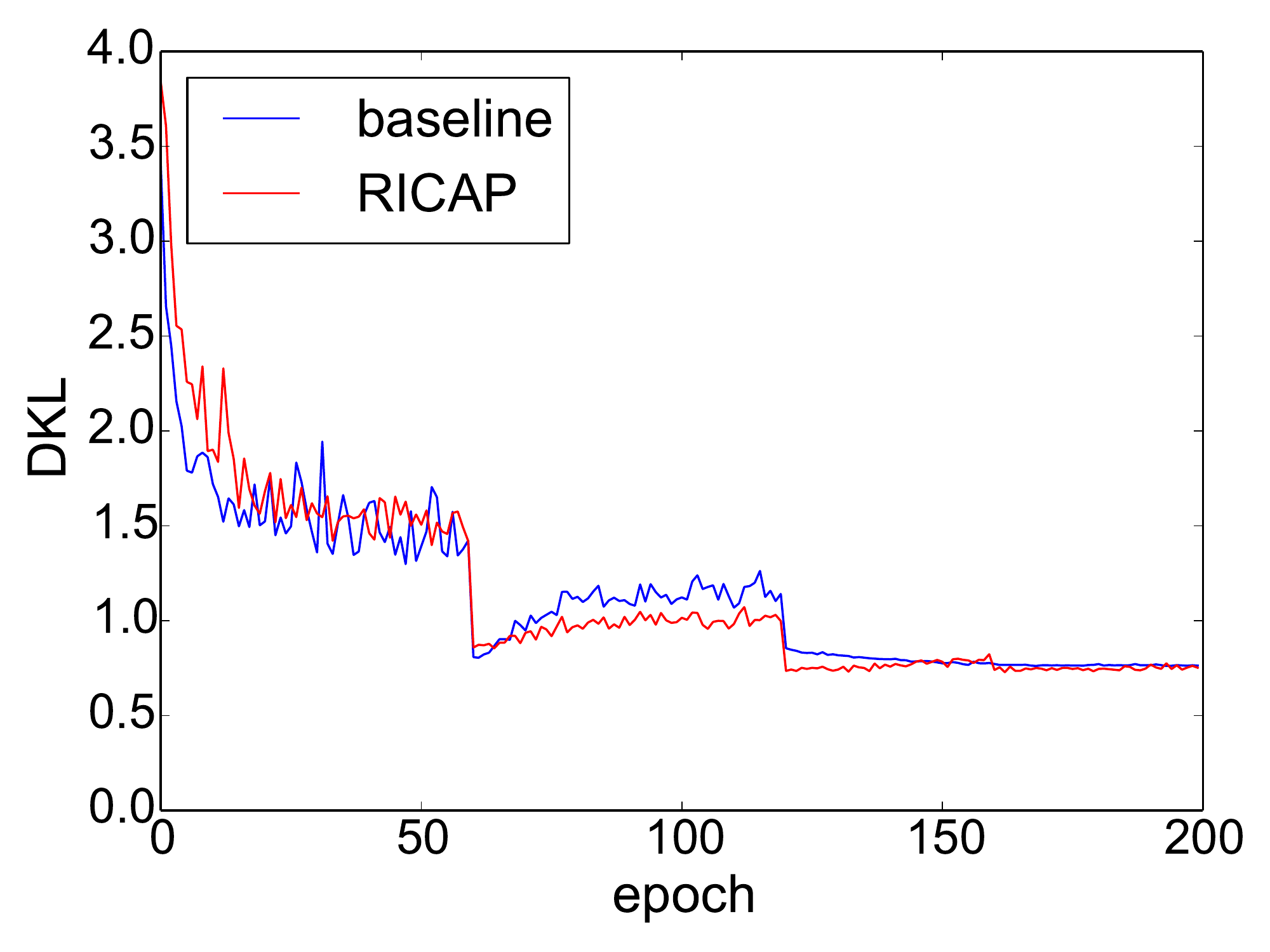}\\
\vspace*{0.4cm}
(e) Test loss on CIFAR-10 &
(f) Test loss on CIFAR-100 \\
\includegraphics[width=2.5in,bb= 0 0 560 420,clip]{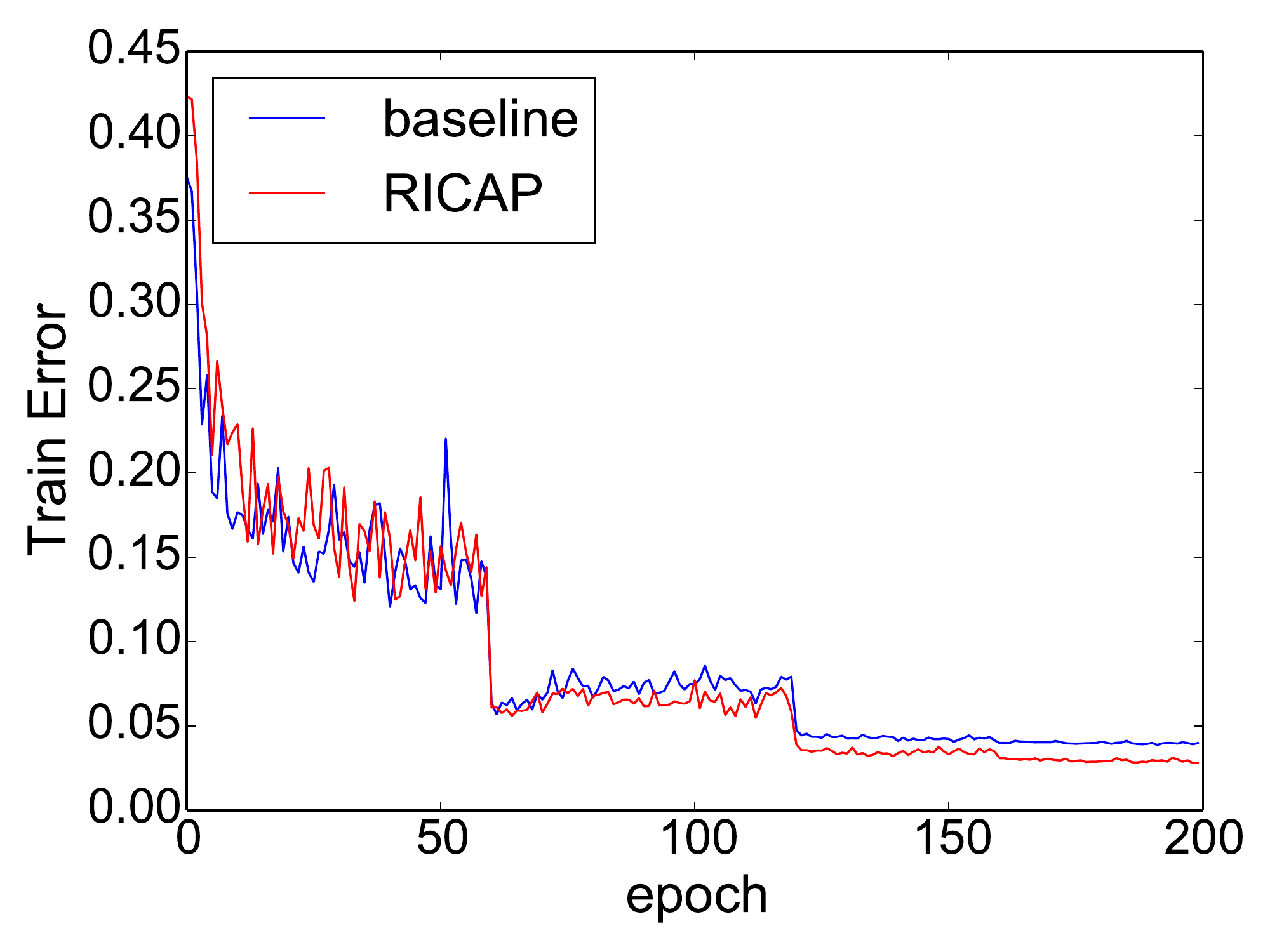}&
\includegraphics[width=2.5in,bb= 0 0 560 420,clip]{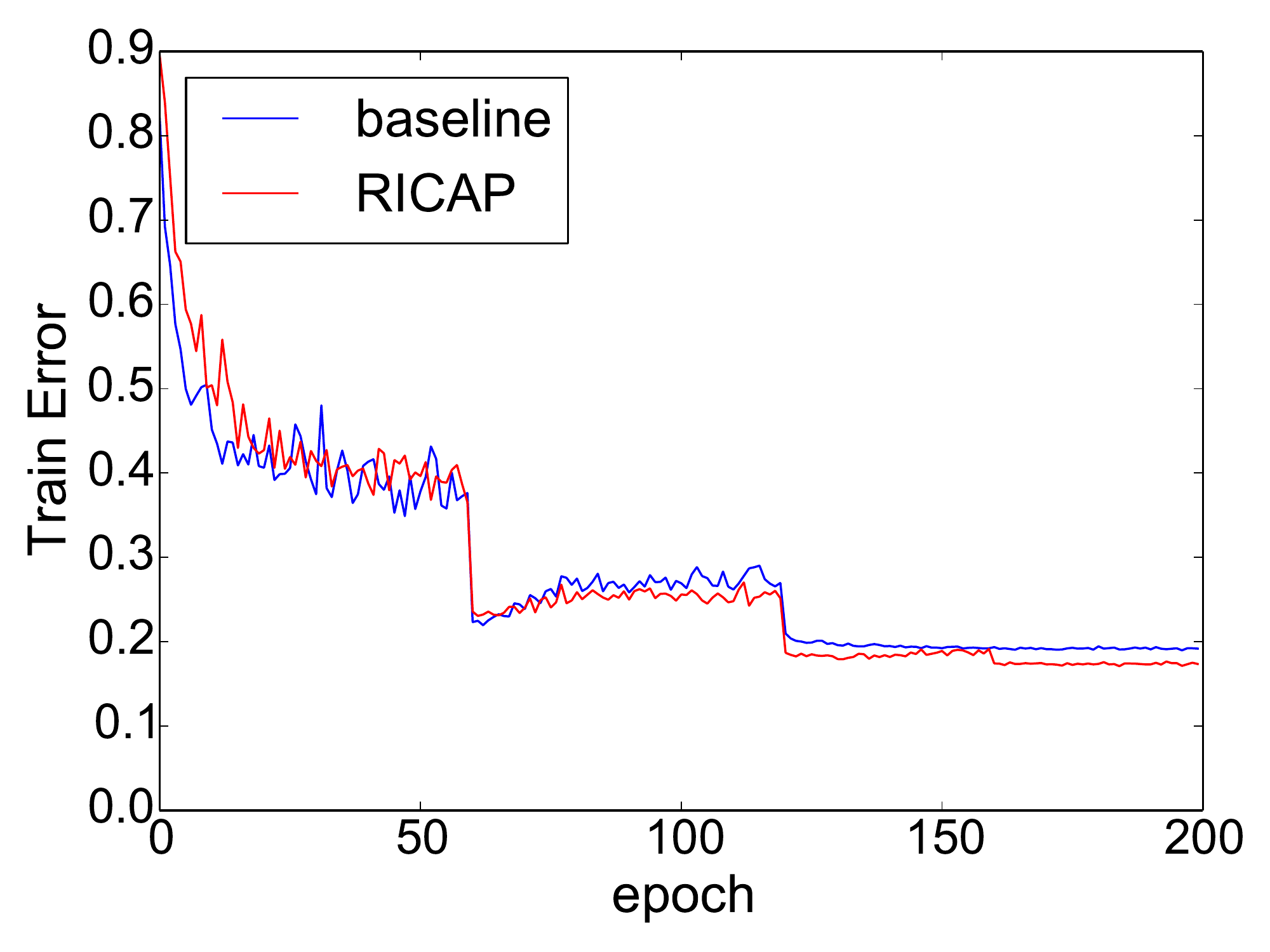}\\
\vspace*{0.4cm}
(g) Test error rate on CIFAR-10 &
(h) Test error rate on CIFAR-100 \\
\end{tabular}
\vspace*{0.5cm}
\caption{
Time-courses of training with and without RICAP.
Note that we plot the Kullback-Leibler divergence as the loss function.
In the case with RICAP, we used the class of the cropped image with the highest occupancy as the correct label to calculate the training error rates for mixed images.
}
\label{fig:ricap_loss}
\end{figure*}

\paragraph*{\textbf{Analysis of Results}}
For further analysis of RICAP, we plotted the losses and error rates in training and test phases with and without RICAP in Fig.~\ref{fig:ricap_loss}.
While the commonly used loss function for multi-class classification is cross-entropy, it converges to zero for hard labels but not for soft labels.
For improving visibility, we employed the Kullback-Leibler divergence as the loss function, which provides the gradients same as the cross-entropy loss and converges to zero for both hard and soft labels.
Also, we used the class of the cropped image with the highest occupancy as the correct label to calculate the training error rates for mixed images.

According to Figs.~\ref{fig:ricap_loss} (a)--(d), training losses and error rates of WideResNet with RICAP never converge to zero and the WideResNet continues to train.
This result implies that RICAP makes the classification training more complicated and hard-to-overfit by image and label mixing.
Figs.~\ref{fig:ricap_loss} (e) and (f) show the test losses of WideResNet at almost the same level with and without RICAP while the test error rates with RICAP in Figs.~\ref{fig:ricap_loss} (g) and (h) are better than baseline.
This is because RICAP prevents the endless pursuit of hard probabilities (which could provide the zero training loss) and overfitting.

\begin{table}[!t]
  \renewcommand{\arraystretch}{1.5}
  \caption{Test Error Rates using WideResNet on the CIFAR dataset.}
  \label{tab:RICAP_WideResNet}
  \centering
  \begin{tabular}{lll}
    \toprule
      Method                       & CIFAR-10                   & CIFAR-100 \\
    \midrule
      Baseline                     & 3.89                       & 18.85                          \\
      + dropout ($p=0.2$)          & 4.65 \(\pm 0.08\)$^\dag$   & 21.27 \(\pm 0.19\)$^\dag$      \\
      + cutout ($16 \times 16$)    & 3.08 \(\pm 0.16\)          & 18.41 \(\pm 0.27\)             \\
      + random erasing             & 3.08 \(\pm 0.05\)          & 17.73 \(\pm 0.15\)             \\
      + mixup ($\alpha=1.0$)       & 3.02 \(\pm 0.04\)$^\dag$   & 17.62 \(\pm 0.25\)$^\dag$      \\
    \midrule
      + RICAP ($\beta=0.3$)        & \textbf{2.85} \(\pm 0.06\) & \textbf{17.22} \(\pm 0.20\)    \\
    \bottomrule
    \multicolumn{3}{l}{$^\dag$ indicates the results of our experiments.}
  \end{tabular}
\end{table}

\begin{table}[!t]
  \renewcommand{\arraystretch}{1.5}
  \caption{Single Crop Test Error Rates of the WideResNet-50-2-bottleneck on ImageNet.}
  \centering
  \label{tab:RICAP_ImageNet_ResNet}
  \begin{tabular}{llll}
    \toprule
      Method                   & Epochs & top-1 Error(\%)             & top-5 Error(\%)  \\
    \midrule
      Baseline                  & 100    & ~~~~ 21.90                  & ~~~~ 6.03             \\
      + cutout ($56 \times 56$) & 100    & ~~~~ 22.45$^\dag$           & ~~~~ 6.22$^\dag$      \\
      + mixup ($\alpha=0.2$)    & 100    & ~~~~ 21.83$^\dag$           & ~~~~ 5.81$^\dag$      \\
      + RICAP ($\beta=0.3$)     & 100    & ~~~~ \textbf{21.08}         & ~~~~ \textbf{5.66}    \\
    \midrule
      Baseline                  & 200    & ~~~~ 21.84$^\dag$           & ~~~~ 6.03$^\dag$      \\
      + cutout ($56 \times 56$) & 200    & ~~~~ 21.51$^\dag$           & ~~~~ 5.89$^\dag$      \\
      + mixup ($\alpha=0.2$)    & 200    & ~~~~ 20.39$^\dag$           & ~~~~ \textbf{5.22}$^\dag$      \\
      + RICAP ($\beta=0.3$)     & 200    & ~~~~ \textbf{20.33}         & ~~~~ 5.26             \\
    \bottomrule
    \multicolumn{3}{l}{$^\dag$ indicates the results of our experiments.}
  \end{tabular}
  \\[1mm]
\end{table}

\begin{table*}[!t]
\renewcommand{\arraystretch}{1.5}
\caption{Test Error Rates on CIFAR-10.}{}
\label{tab:RICAP_Others}
\centering
  \begin{tabular}{llll}
    \toprule
      Method                           & DenseNet-BC 190-40                   & Pyramidal ResNet 272-200             & Shake-Shake 26 2x96d                 \\
    \midrule
      Baseline                         & ~~~~~~~~ 3.46                       & ~~~~~~~~ 3.31 \(\pm 0.08\)           & ~~~~~~~~ 2.86                       \\
      + dropout ($p=0.2$)              & ~~~~~~~~ 4.56 $^\dag$               & ~~~~~~~~ 4.06 $^\dag$                & ~~~~~~~~ 3.79 $^\dag$               \\
      + cutout ($8 \times 8$)          & ~~~~~~~~ 2.73 \(\pm 0.06\)$^\dag$   & ~~~~~~~~ 2.84 \(\pm 0.05\)$^\dag$    & ~~~~~~~~ 2.56 \(\pm 0.07\)          \\
      + mixup ($\alpha=1.0$)           & ~~~~~~~~ 2.73 \(\pm 0.08\)$^\dag$   & ~~~~~~~~ 2.57 \(\pm 0.09\)$^\dag$    & ~~~~~~~~ 2.32 \(\pm 0.11\)$^\dag$   \\
    \midrule
      + RICAP ($\beta=0.3$)            & ~~~~~~~~ \textbf{2.69} \(\pm 0.12\) & ~~~~~~~~  \textbf{2.51} \(\pm 0.02\) & ~~~~~~~~ \textbf{2.19} \(\pm 0.08\) \\
    \bottomrule
    \multicolumn{3}{l}{$^\dag$ indicates the results of our experiments.}
  \end{tabular}
\end{table*}

\indent
\subsection{Classification of ImageNet}
\label{sec:imagenet}

In this section, we evaluate RICAP on the classification task of the ImageNet dataset~\cite{Russakovsky2014}.
ImageNet consists of 1.28 million training images and 50,000 validation images.
Each image is given one of 1,000 class labels.
We normalized each channel of all images to zero mean and unit variance as preprocessing.
We also employed random resizing, random $224\times224$ cropping, color jitter, lighting, and random flipping in the horizontal direction following previous studies~\cite{Zagoruyko2016,Zhang2017}.

To evaluate RICAP, we applied it to the \emph{WideResNet 50-2-bottleneck} architecture, consisting of $50$ convolution layers using bottleneck residual blocks with a widen factor of $2$ and dropout with a drop probability of $p=0.3$ in intermediate layers~\cite{Zagoruyko2016}.
This architecture achieved the highest accuracy on ImageNet in the original study~\cite{Zagoruyko2016}.
The hyperparameters and other conditions were the same as those used in the baseline study.
WideResNet 50-2-bottleneck was trained using the momentum SGD algorithm with a momentum parameter of 0.9 and weight decay of $10^{-4}$ over $100$ or $200$ epochs with batches of $256$ images.
The learning rate was initialized to $0.1$, and then, it was reduced to $0.01$, $0.001$ and $0.0001$ at the $30$th, $60$th, and $90$th-epoch, respectively, in the case of $100$ epoch training.
The learning rate was reduced at the $65$th, $130$th, and $190$th-epoch, respectively, in the case of $200$ epoch training.
For our RICAP, we used the hyperparameter $\beta=0.3$ according to the results of Section.~\ref{sec:c10_c100}.

Table~\ref{tab:RICAP_ImageNet_ResNet} summarizes the results of RICAP with the WideResNet 50-2-bottleneck as well as the results of competitive methods: cutout~\cite{DeVries2017} and mixup~\cite{Zhang2017}.
Competitive results denoted by $\dag$ symbols were obtained from our experiments and the other results are cited from the original studies.
We used $\alpha=0.2$ for mixup according to the original study.
Cutout did not attempt to apply cutout to the ImageNet dataset.
It used a cutout size of $8\times8$ for the CIFAR-10, in which an image has a size of $32\times32$.
Since a preprocessed image in the ImageNet dataset has a size of $224\times224$, we multiplied the cutout size by 7 ($224/32$) to apply cutout to the ImageNet dataset.

RICAP clearly outperformed the baseline and competitive methods in the case of $100$ epoch training, and was superior or competitive to the others in the case of $200$ epoch training.
Compared to RICAP, cutout and mixup require a longer training to get results better than the baseline.
This is because, as mentioned in Section~\ref{sec:concept}, cutout reduces the amount of available features in each and mixup generates pixel-level features that original images never produce.

Mixup requires careful adjustment of the hyperparameter; the best hyperparameter value is $\alpha=1.0$ for the CIFAR datasets and $\alpha=0.2$ for the ImageNet dataset.
An inappropriate hyperparameter reduces performance significantly~\cite{Zhang2017}.
On the other hand, RICAP with the hyperparameter $\beta=0.3$ achieved significant results in both the CIFAR and ImageNet datasets.
Furthermore, the bottom panels in Fig.~\ref{fig:fusion-opt} show the robustness of RICAP to the hyperparameter value.

\indent
\subsection{Classification by Other Architectures}
\label{sec:others}
We also evaluated RICAP with DenseNet~\cite{Huang2016b}, the pyramidal ResNet~\cite{Han2016}, and the shake-shake regularization model~\cite{Gastaldi2017a} on the CIFAR-10 dataset~\cite{Krizhevsky2009}.
For the DenseNet, we used the architecture \emph{DenseNetBC 190-40}; as the name implies, it consists of $190$ convolution layers using bottleneck residual blocks with a growing rate of $40$.
For the pyramidal ResNet, we used the architecture \emph{Pyramidal ResNet 272-200}, which consists of $272$ convolution layers using bottleneck residual blocks with a widening factor of $200$.
For the shake-shake regularization model, we used the architecture \emph{ShakeShake 26 2$\times$96d}; this is a ResNet with $26$ convolution layers and $2\times96$d channels with shake-shake image regularization.
These architectures achieved the best results in the original studies.
We applied data normalization and data augmentation in the same way as Section.~\ref{sec:c10_c100}.
The hyperparameters were the same as those in the original studies~\cite{Huang2016b,Han2016,Gastaldi2017a}.

We summarized the results in Table~\ref{tab:RICAP_Others}.
We used the hyperparameter $\beta=0.3$ according to the results of Section.~\ref{sec:c10_c100}.
RICAP outperformed the competitive methods.
In particular, the shake-shake regularization model with RICAP achieved a test error rate of $2.19\%$; this is a new record on the CIFAR-10 classification among the studies under the same conditions~\cite{Huang2016b,Han2016,Gastaldi2017a,DeVries2017,Zhong2017a,Zhang2017}\footnote{AutoAugment~\cite{Cubuk2018} achieved a further improved result by employing additional data augmentation techniques such as shearing and adjusting their parameters.}.
These results also indicate that RICAP is applicable to various CNN architectures and the appropriate hyperparameter does not depend on the CNN architectures.

\begin{figure*}[!t]{}
\centering
\includegraphics[width=7.5in,bb= 0 150 840 450,clip]{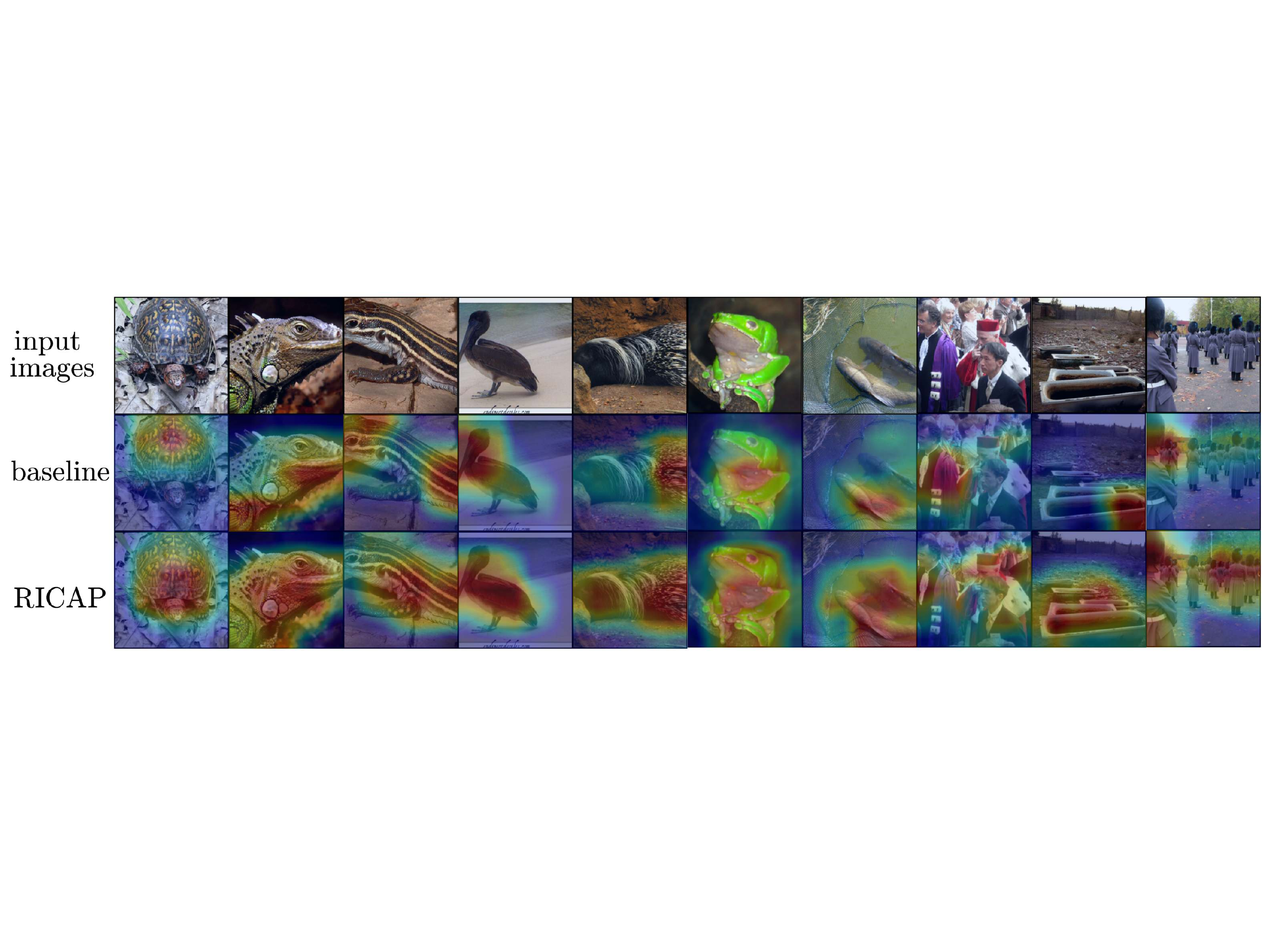}
\vspace*{-12mm}
\caption{
Class Activation Mapping (CAM)~\cite{Zhou2016} of WideResNet 28-10.
The top row shows the input images.
The middle row shows the CAM of WideResNet 28-10 without RICAP denoted as \emph{baseline}.
The bottom row shows the CAM of WideResNet 28-10 with RICAP.
}
\label{fig:ricap_cam}
\end{figure*}
\vspace{1cm}


\section{Visualization and Qualitative Analysis in Classification}
\label{sec:analysis}

In this section we analyze thg effectiveness of RICAP in detail through the visualization, ablation study and comparison with other data augmentation methods.

\subsection{Visualization of Feature Learning by RICAP}
\label{sec:visualization}

One of the most serious overfitting of a CNN arises when classifying images according a limited set of features and ignoring others.
For example, if a CNN classifies cat images according to features of the cats' face, it fails to classify an image that depicts a cats' back.
Since RICAP collects and crops four images randomly, each image provides a different cropped region in every training step.
This is expected to support the CNN in using a wider variety of features from the same image and to prevent the CNN from overfitting to features of a specific region.

To verify this hypothesis, we visualized the regions in which a CNN focuses much attention using the \emph{Class Activation Mapping (CAM)}~\cite{Zhou2016}.
The CAM expects a CNN to have a global average pooling layer to obtain the spatial average of the feature map before the final output layer.
The CAM calculates the regional importance by projecting back the output (typically, the correct label) to the feature map.

Fig.~\ref{fig:ricap_cam} shows the CAMs of the WideResNet 50-2-bottleneck with and without RICAP.
This model was trained in the previous ImageNet experiments in Section~\ref{sec:imagenet}.
The top row shows the input images.
The middle row denoted as the baseline shows the CAMs of WideResNet without RICAP.
WideResNet focuses attention on limited regions of objects in the first to sixth columns: the shell of a turtle and the faces of animals.
WideResNet focuses attention on objects in the foreground and ignores objects in the background in the seventh and tenth columns.
The bottom row shows the CAMs of WideResNet with RICAP.
WideResNet focuses attention on the whole bodies of animals in the first to sixth columns and objects in the foreground and background in the seventh to tenth columns.
These results demonstrate that RICAP prevents the CNN from overfitting to specific features.

\begin{figure*}[!t]{}
\vspace*{5mm}
\centering
\includegraphics[width=5.5in,bb= 0 0 860 600,clip]{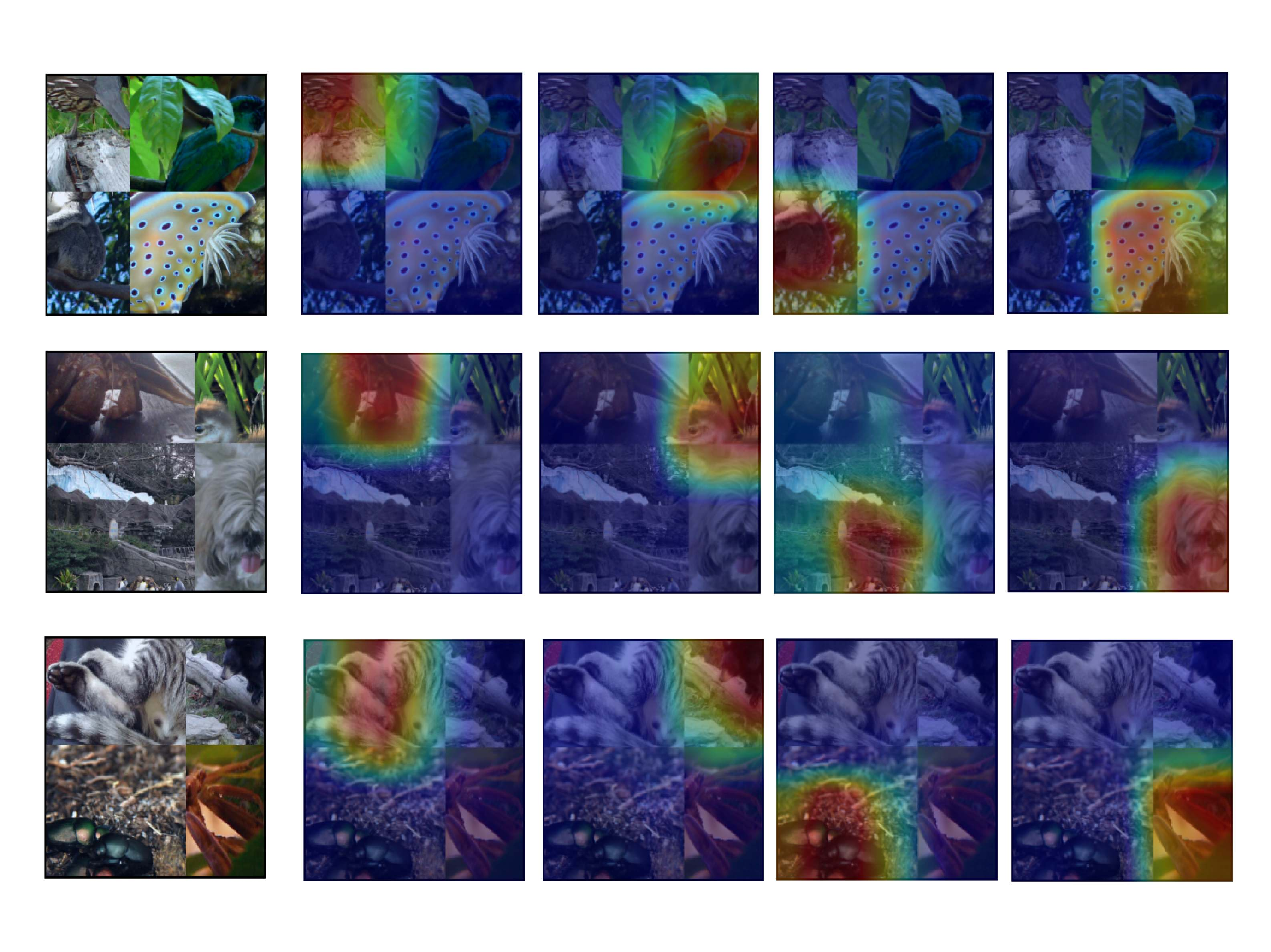}
\vspace*{-5mm}
\caption{
Class Activation Mapping (CAM)~\cite{Zhou2016} of WideResNet 28-10 using the images that RICAP cropped and patched.
The leftmost column shows the input images.
The second to fifth columns show the CAMs by projecting back the labels corresponding to the upper left, upper right, lower left, and lower right image patches, respectively.
}
\label{fig:ricap_cam2}
\end{figure*}

In addition, we visualized the CAMs of the WideResNet using the images that RICAP cropped and patched in Fig.~\ref{fig:ricap_cam2}.
The leftmost column shows the input images.
The second to fifth columns show the CAMs obtained by projecting back the labels corresponding to the upper left, upper right, lower left, and lower right image patches, respectively.
The CAMs demonstrated that WideResNet focuses attention on the object corresponding to each given label correctly, even though the depicted objects were extremely cropped.
Moreover, we confirmed that WideResNet automatically learns to ignore the boundary patching caused by RICAP and potentially becomes robust to occlusion and cutting off.

\begin{figure*}[!t]{}
\vspace*{5mm}
\centering
\includegraphics[width=7.0in,bb= 0 70 860 530,clip]{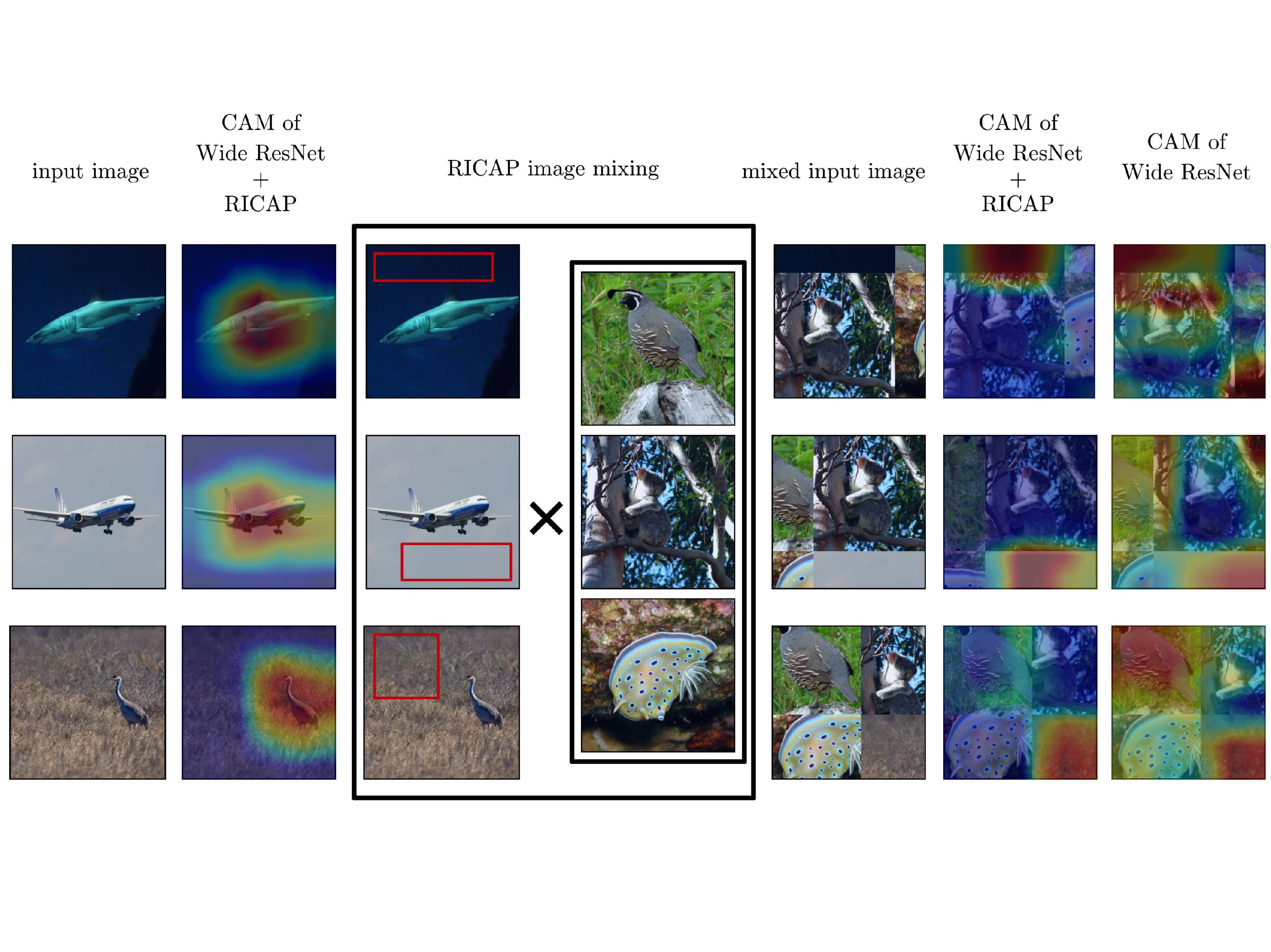}
\caption{
Class Activation Mapping (CAM)~\cite{Zhou2016} when only a background area is cropped and patched by RICAP.
(leftmost column) We randomly chose $3$ base images depicting objects in long shots.
(second column) The CAMs confirm that the WideResNet trained with RICAP pays much attention to the objects.
(third--fifth columns) The backgrounds in the base images are cropped and patched with $3$ other randomly chosen images.
(two rightmost columns)
The CAMs obtained from the WideResNets trained with and without RICAP.
}
\label{fig:background}
\end{figure*}

\indent
\subsection{Case with No Objects in Cropped Areas}
\label{sec:visualization_background}
RICAP does not check whether an object is in a cropped area.
In this section, we evaluate the case that the cropped region contains no object.
We prepared two WideResNets trained with and without RICAP.
We randomly chose $3$ images depicting objects in long shots as shown in the first column of Fig.~\ref{fig:background}.
We can confirm that the WideResNet trained with RICAP pays much attention to the objects using \emph{Class Activation Mapping (CAM)}~\cite{Zhou2016} as shown in the second column.
We cropped only the backgrounds from the former $3$ images and patched with other randomly chosen $3$ images using RICAP algorithm, obtaining $3$ input images, as shown in third to fifth columns.
Then, we fed the $3$ input images to WideResNets trained with and without RICAP and obtained the CAMs for the classes of the background images in the two rightmost columns.
The CAMs show that the WideResNet trained with RICAP focuses only on the cropped background images whereas the WideResNet trained without RICAP pays attention to many subregions loosely.
They demonstrate that RICAP enabled the WideResNet to learn even the features of backgrounds as clues to classifying images.
Compared with the case of the non-cropped images in the second column, we can conclude that RICAP enables a CNN to pay attention to the objects if they clearly exist and to other clues such as backgrounds otherwise.

\indent
\section{Ablation Study and Detailed Comparison}
\label{sec:detailed}
\subsection{Ablation Study}
\label{sec:ablation}

RICAP is composed of two manipulations: image mixing (cropping and patching) and label mixing.
For image mixing, RICAP randomly selects, crops, and patches four images to construct a new training image.
For label mixing, RICAP mixes class labels with ratios proportional to the areas of four images.
In this section, we evaluated the contributions of these two manipulations using \emph{WideResNet 28-10} and the CIFAR-10 and CIFAR-100 datasets~\cite{Krizhevsky2009} as in Section.~\ref{sec:c10_c100}.
Data normalization, data augmentation, and the hyperparameters were also the same as those used in Section.~\ref{sec:c10_c100}.
We chose the hyperparameter $\beta$ of RICAP from $\{0.1, 0.3, 1.0\}$.
The results are summarized in Table~\ref{tab:ablation}.

First, we performed image mixing without label mixing.
We used the class label of the patched image that had the largest area as the target class label, i.e., $c=c_k$ for $k=\arg\max_{k'\in\{1,2,3,4\}} W_{k'}$ instead of Eq.~\eqref{eq:mixweight}.
Using only image mixing, WideResNet achieved much better results than the baseline but was not competitive in the case with label mixing.

Second, we performed label mixing without image mixing.
In this case, we used the boundary position to calculate only the ratio of label mixing and we used the original image with the largest probability as the training sample.
WideResNet achieved much worse results, demonstrating the harmful influence of extreme soft labeling.

We conclude that both image and label mixing jointly play an important role in RICAP.

\begin{table}[!t]
  \renewcommand{\arraystretch}{1.5}
  \caption{Test Error Rates using WideResNet in the Ablation Study.}
  \label{tab:ablation}
  \centering
  \begin{tabular}{lll}
    \toprule
      Method                      & CIFAR-10                   & CIFAR-100                      \\
    \midrule
      Baseline                    & 3.89                       & 18.85                          \\
      + mixup ($\alpha=1.0$)      & 3.02 \(\pm 0.04\)$^\dag$   & 17.62 \(\pm 0.25\)$^\dag$      \\
    \midrule
      + RICAP (image mixing only, $\beta=0.1$)       & 3.34 \(\pm 0.09\) & 17.87 \(\pm 0.22\)   \\
      + RICAP (image mixing only, $\beta=0.3$)       & 3.33 \(\pm 0.10\) & 17.95 \(\pm 0.13\)   \\
      + RICAP (image mixing only, $\beta=1.0$)       & 3.70 \(\pm 0.07\) & 18.90 \(\pm 0.24\)   \\
    \midrule
      + RICAP (label mixing only, $\beta=0.1$) & 69.28             & -                          \\
      + RICAP (label mixing only, $\beta=0.3$) & 62.84             & -                          \\
      + RICAP (label mixing only, $\beta=1.0$) & 68.91             & -                          \\
    \midrule
      + $4$ mixup ($\beta=0.1$)   & 3.29 \(\pm 0.07\)$^\dag$       & 17.62 \(\pm 0.21\)$^\dag$  \\
      + $4$ mixup ($\beta=0.3$)   & 3.11 \(\pm 0.05\)$^\dag$       & 18.04 \(\pm 0.16\)$^\dag$  \\
      + $4$ mixup ($\beta=1.0$)   & 3.71 \(\pm 0.17\)$^\dag$       & 19.57 \(\pm 0.15\)$^\dag$  \\
    \midrule
      + RICAP ($\beta=0.1$)       & 3.01 \(\pm 0.15\)          & 17.39 \(\pm 0.09\)             \\
      + RICAP ($\beta=0.3$)       & \textbf{2.85} \(\pm 0.06\) & \textbf{17.22} \(\pm 0.20\)    \\
      + RICAP ($\beta=1.0$)       & 2.91 \(\pm 0.01\)          & 17.82 \(\pm 0.03\)             \\
    \bottomrule
    \multicolumn{3}{l}{$^\dag$ indicates the results of our experiments.}
  \end{tabular}
\end{table}

\indent
\subsection{Comparison with Mixup of four Images}
\label{sec:4mixup}

RICAP patches four images spatially and mixes class labels using the areas of the patched images.
One can find a similarity between RICAP and mixup; mixup alpha-blends two images and mixes their class labels using the alpha value.
The main difference between these two methods is that between spatially patching and alpha-blending.
Another difference is the number of mixed images: four for RICAP and two for mixup.

Here, as a simple extension of mixup, we evaluated mixup that mixes four images and call it $4$-mixup.
In this experiment, we used the \emph{WideResNet 28-10} and the CIFAR-10 and CIFAR-100 datasets~\cite{Krizhevsky2009} as in Section.~\ref{sec:c10_c100}.
Data normalization, data augmentation, and hyperparameters were also the same as those used in Section.~\ref{sec:c10_c100}.
The alpha values were chosen in the same way as RICAP with the hyperparameter $\beta$.

We summarized the results in Table~\ref{tab:ablation}.
While $4$-mixup had better results than the baseline, it had worse results than both the original mixup and RICAP.
Increasing the number of images cannot improve the performance of mixup.
This suggests that RICAP owes its high performance not to the number of images or to the ability to utilize four images.


\section{Experiments on Other Tasks}
\label{sec:other_task}

In this section, we evaluate RICAP on image-caption retrieval, person re-identification, and object detection to demonstrate the generality of RICAP.

\indent
\subsection{Evaluation on Image-Caption Retrieval}
\label{sec:image-caption}
\paragraph*{\textbf{Image-Caption Retrieval}}

For image-caption retrieval, the main goal is to retrieve the most relevant image for a given caption and to retrieve the most relevant caption for a given image.
A dataset contains pairs of images $i_n$ and captions $c_n$.
An image $i_n$ is considered the most relevant to the paired caption $c_n$ and vice versa.
A relevant pair $(i_n,c_n)$ is called positive, and an irrelevant pair $(i_n,c_m)$ ($m\neq n$) is called negative.
The performance is often evaluated using recall at $K$ (denoted as $R@K$) and $Med\ r$.

A common approach for image-caption retrieval is called \emph{visual-semantic embeddings (VSE)}~\cite{Kiros2014}.
Typically, a CNN encodes an image into a vector representation and a recurrent neural network (RNN) encodes a caption to another vector representation.
The neural networks are jointly trained to build a similarity function that gives higher scores to positive pairs than negative pairs.
\emph{VSE++}~\cite{Faghri2017} employed a ResNet$152$~\cite{He2016} as the image encoder and a GRU~\cite{Kiros2014} as the caption encoder and achieved remarkable results.
We used VSE++ as a baseline of our experiments.

First, we introduce the case without RICAP.
An image $i_n$ is fed to the image encoder $ResNet(\cdot)$ and encoded to a representation $v_{i_n}$ as
\begin{equation*}
  v_{i_n} = ResNet(i_n).
\end{equation*}
A caption $c_n$ is fed to the caption encoder $GRU(\cdot)$ and encoded to a representation $v_{c_n}$ as
\begin{equation*}
  v_{c_n} = GRU(c_n).
\end{equation*}
VSE++ $S(\cdot,\cdot)$ defines the similarity $S_n$ between the pair $(i_n, c_n)$ as
\begin{align*}
  S_n & = S(v_{i_n},v_{c_n}), \\
      & = S(ResNet(i_n),GRU(c_n)).
\end{align*}
Refer to the original study~\cite{Faghri2017} for a more detailed implementation.

\begin{table*}[!t]
\vspace*{5mm}
\renewcommand{\arraystretch}{1.5}
\caption{Results of Image-Caption Retrieval using Microsoft COCO.}
\label{tab:RICAP_Image_Caption_Retrieval_VSE}
\centering
  \begin{tabular}{llllllllll}
    \toprule
      Model & \multicolumn{4}{c}{Caption Retrieval} && \multicolumn{4}{c}{Image Retrieval} \\
      \cline{2-5}\cline{7-10}
      & R@1 & R@5 & R@10 & Med r && R@1 & R@5 & R@10 & Med r \\
    \midrule
      Baseline                         & 64.6 & 90.0 & 95.7 & \textbf{1.0} && 52.0 & 84.3 & 92.0 & \textbf{1.0} \\
    \midrule
      + RICAP ($\beta=0.3$)            & \textbf{65.8} & \textbf{90.2} & \textbf{96.2} & \textbf{1.0} && \textbf{52.3} & \textbf{84.4} & \textbf{92.4} & \textbf{1.0} \\
    \bottomrule
  \end{tabular}
\vspace*{5mm}
\end{table*}

\paragraph*{\textbf{Application RICAP to VSE++}}

With RICAP, we propose a new training procedure for the image encoder.
We randomly selected four images $i_m$, $i_n$, $i_o$, and $i_p$ and created a new image $i_{ricap}$ as the cropping and patching procedure in Section.~\ref{sec:proposed_method}.
\begin{equation*}
  i_{ricap} = RICAP_{image}(i_m,i_n,i_o,i_p).
\end{equation*}
The function \emph{$RICAP_{image}(\cdot,\cdot,\cdot,\cdot)$} denotes the procedure of RICAP proposed in Section~\ref{sec:proposed_method_general}.
The image encoder $ResNet(\cdot)$ encodes the patched image $i_{ricap}$ to a representation $v_{i_{ricap}}$ as
\begin{equation*}
  v_{i_{ricap}} = ResNet(i_{ricap}).
\end{equation*}
As a paired caption representation, we obtained the average of the relevant caption representations.
The specific procedure was as follows.
We fed the paired captions $c_m$, $c_n$, $c_o$, and $c_p$ individually to the caption encoder and encoded them into the representations $v_{c_m}$, $v_{c_n}$, $v_{c_o}$, and $v_{c_p}$, respectively.
Next, we averaged the caption representations $v_{c_m}$, $v_{c_n}$, $v_{c_o}$, and $v_{c_p}$ with ratios proportional to the areas of the cropped images and obtained the mixed vector $v_{c_{ricap}}$ as
\begin{align*}
  v_{c_{ricap}}
  &=RICAP_{caption}(v_{c_m},v_{c_n},v_{c_o},v_{c_p})\\
  &\textstyle:=\sum_{k=\{m,n,o,p\}}W_k v_{c_k},\\
  &\textstyle=\sum_{k=\{m,n,o,p\}}W_k GRU(c_k),
\end{align*}
where $W_k$ is the area ratio of the image $k$ as in Eq.~\eqref{eq:mixweight}.
Here, we used the vector representation $v_{c_{ricap}}$ as the one positively paired with the vector representation $v_{i_{ricap}}$ and obtained the similarity $S_{ricap}$ between this pair.
In short, we used the following similarity to train the image encoder;
\begin{align*}
  & S_{ricap}\\
  &\ = S(v_{i_{ricap}},v_{c_{ricap}}), \\
  &\ = S(ResNet(RICAP_{image}(i_m,i_n,i_o,i_p)), \\
  &\ \ \ \  RICAP_{\!caption}\!(GRU\!(c_m)\!,GRU\!(c_n)\!,GRU\!(c_o)\!,GRU\!(c_p))). \\
\end{align*}
We treated the remaining vector representations $v_{c_k}$ for $k\notin\{m,n,o,p\}$ as negative pairs.
Note that we used the ordinary similarity $S_n$ to train the caption encoder.

\paragraph*{\textbf{Experiments and Results}}

We used the same experimental settings as in the original study of VSE++~\cite{Faghri2017}.
We used the Microsoft COCO dataset~\cite{Lin2014}; $113,287$ images for training model and $1,000$ images for validation.
We summarized the score averaged over $5$ folds of $1,000$ test images.

The ResNet was pre-trained using the ImageNet dataset, and the final layer was replaced with a new fully-connected layer.
For the first 30 epochs, the layers in the ResNet except for the final layer were fixed.
The GRU and the final layer of the ResNet were updated using the Adam optimizer~\cite{Kingma2014} using a mini-batch size of $128$ with the hyperparameter $\alpha=0.0002$ for the first $15$ epochs and then $\alpha=0.00002$ for the other $15$ epochs.
Next, the whole model was fine-tuned for the additional $15$ epochs with $\alpha=0.00002$.

Table~\ref{tab:RICAP_Image_Caption_Retrieval_VSE} summarizes the results; RICAP improved the performance of VSE++.
This result demonstrated that RICAP is directly applicable to image processing tasks other than classification.

\indent
\subsection{Evaluation on Person Re-identification}
\label{sec:person_reid}
\begin{figure}[!t]{}
  \vspace*{5mm}
  \centering
  \includegraphics[width=3.6in,bb= 100 30 760 550,clip]{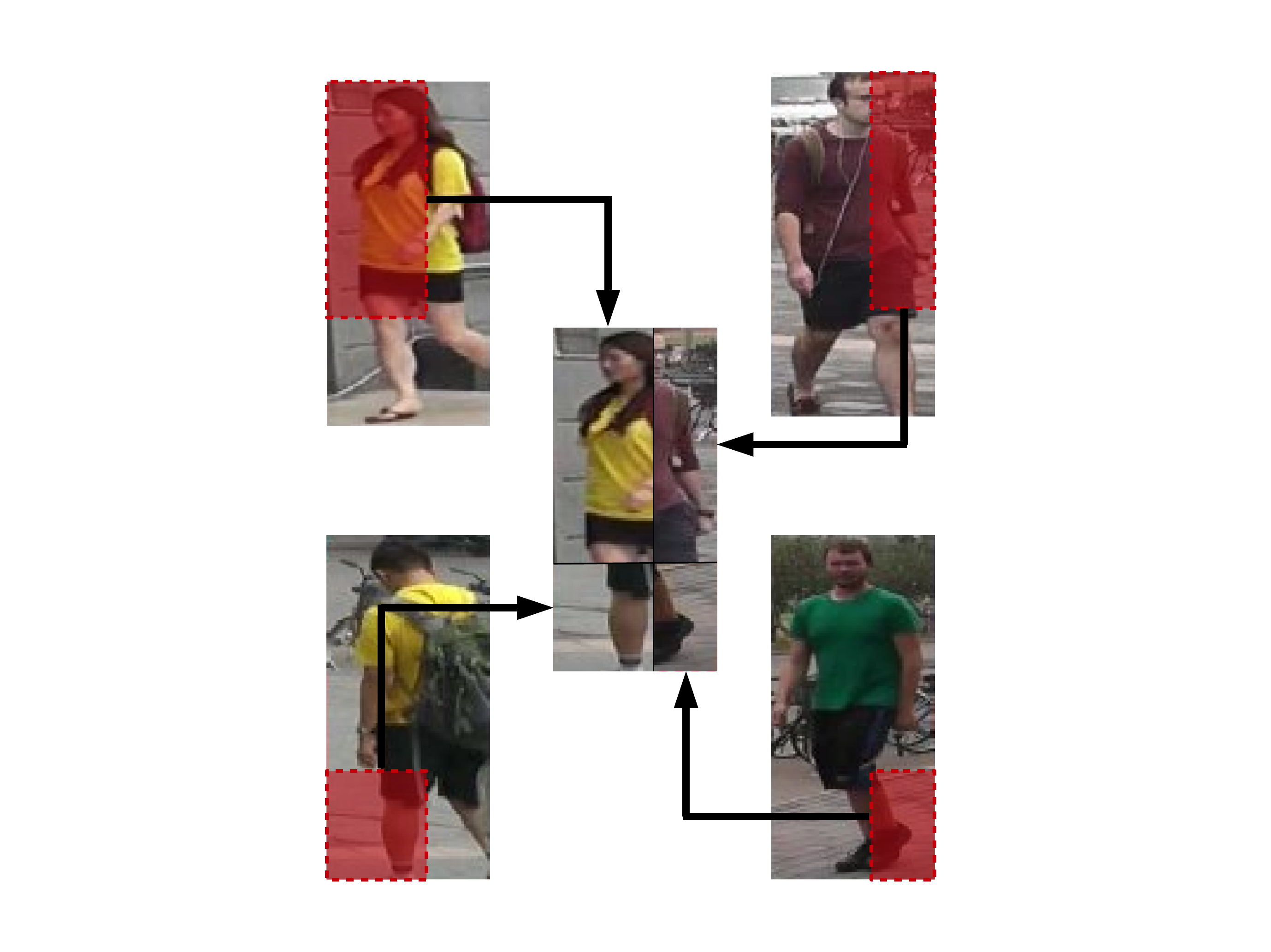}
  \caption{
  Conceptual explanation of the proposed \emph{fixed image cropping and patching} (\emph{FICAP}) data augmentation.
  For application of RICAP to person re-identification task, we replaced the random cropping with fixed cropping because of the feature vector matching for image-to-image retrieval.
  }
  \label{fig:ficap}
\end{figure}

\paragraph*{\textbf{Person Re-identification}}
In this section, we evaluate RICAP with a person re-identification task.
For person re-identification, the main goal is to retrieve images of the person identical to a person in a given image.
Deep learning methods train a classifier of persons with IDs and retrieve persons based on the similarity in internal feature vector.
ID discriminative embedding (IDE)~\cite{Zheng2017a} is commonly used baseline consisting of deep feature extractor and classifier.
The performance is often evaluated using recall at K (denoted as R@K) and mean average precision (denoted as mAP).
RICAP also handles partial features and we evaluate RICAP with the IDE.
\paragraph*{\textbf{Application RICAP to IDE}}
Unlike natural images such as images in CIFAR and ImageNet datasets, typical images for person re-identification are already aligned and cropped to depict persons at the center.
Hence, the absolute positions of human parts in images are meaningful for retrieval.
To adopt RICAP to this situation, we crop each image not randomly but considering its absolute position, that is $x_k$ is fixed to $0$ for $k\in\{1,3\}$ and $w$ for $k\in\{2,4\}$, and $y_k$ is fixed $0$ for $k\in\{1,2\}$ and $h$ for $k\in\{3,4\}$.
We call this modified RICAP as \emph{fixed image cropping and patching (FICAP)} and show the overview in above of Fig.~\ref{fig:ficap}.
Note that the boundary position is still determined randomly.

\begin{table}[!t]
  \vspace*{5mm}
  \renewcommand{\arraystretch}{1.5}
\caption{Results of Person Re-identification on Market-1501.}
  \label{tab:reid}
  \centering
  \begin{tabular}{llllll}
    \toprule
      Model & R@1 & R@5 & R@10 & mAP \\
    \midrule
      IDE                        & 85.3 & 94.0 & 96.3 & 68.5 \\
    \midrule
      IDE + FICAP ($\beta=0.1$)  & \textbf{89.8} & \textbf{96.0} & \textbf{97.6} & \textbf{74.6} \\
      IDE + FICAP ($\beta=0.3$)  & 88.4 & 95.0 & 97.0 & 73.2 \\
    \bottomrule
  \end{tabular}
  \vspace*{5mm}
\end{table}

\paragraph*{\textbf{Experiments and Results}}
We used the IDE with the experimental setting introduced as a strong baseline in the PCB paper~\cite{Yifan2018}.
A pixel intensity of each channel was normalized as a preprocessing.
As a data augmentation, training images were randomly flipped in the horizontal direction.
The weight parameters were updated using the momentum SGD algorithm with a momentum parameter of 0.9 and weight decay of $10^{-4}$ over $60$ epochs with batches of $64$ images.
The learning rate was initialized to $0.1$, and then, it was reduced to $0.01$ at the $40$th epochs.
The backbone model was the $50$ layer ResNet~\cite{He2016} pre-trained on ImageNet.
We used the Market1501 dataset~\cite{Zheng2015b}, consisting of $32,668$ images of $1,501$ identities shoot by $6$ cameras.
$12,936$ images of $751$ identities were used for training and $19,732$ images of $750$ identities were used for test.
Table~\ref{tab:reid} summarizes the experimental results, where we cited the results of IDE from the PCB paper~\cite{Yifan2018} for fair comparison.
The result demonstrates that FICAP improved the identification performance of the IDE, indicating the generality of FICAP.

As a further analysis, we implemented FICAP on the state-of-the-art method, PCB, as summarized in the Table~\ref{tab:reid_pcb}.
We employed PCB + RPP model, which is the model of the highest performance in the PCB paper~\cite{Yifan2018}.
PCB divides an image into six subparts horizontally and evaluates the similarity summed over the subparts.
We applied FICAP to each subpart, resulting in 24 patches per person, and found that the performance was degraded.
The same went for the case that the number of patches per subpart was reduced to two as summarized in the Table~\ref{tab:reid_pcb}.
This can be because the image cropping and patching by FICAP conflicts with the image division by PCB and each patch becomes too small to be recognized.
While RICAP and FICAP are general approaches, they are not compatible with methods which already divide images into many subparts.


\begin{table}[!t]
\vspace*{5mm}
\renewcommand{\arraystretch}{1.5}
\caption{Results of Person Re-identification on Market-1501.}
\label{tab:reid_pcb}
\centering
  \begin{tabular}{llllll}
    \toprule
      Model & R@1 & R@5 & R@10 & mAP \\
    \midrule
      PCB + RPP                        & \textbf{93.8} & \textbf{97.5} & \textbf{98.5} & \textbf{81.6} \\
    \midrule
      PCB + RPP + FICAP ($\beta=0.1$)  & 93.0 & 97.4 & 98.5 & 81.4 \\
      PCB + RPP + FICAP ($\beta=0.3$)  & 92.1 & 97.0 & 98.2 & 80.5 \\
    \bottomrule
  \end{tabular}
\vspace*{5mm}
\end{table}

\begin{figure*}[!t]{}
\centering
\includegraphics[width=7.5in,bb= 0 30 840 450,clip]{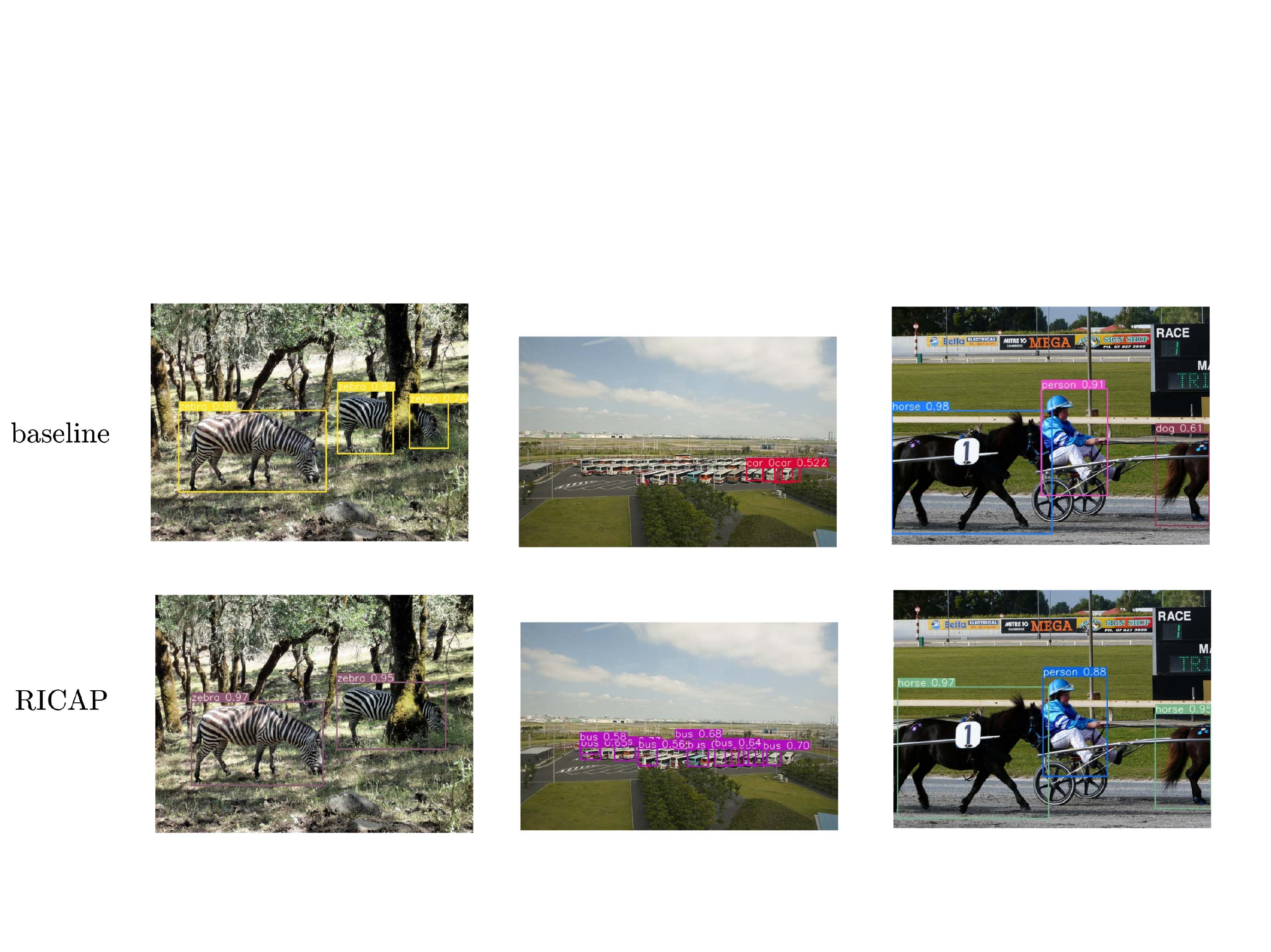}
\vspace*{-12mm}
\caption{
Detection examples of MS-COCO test images by the baseline YOLOv3 (top row) and a TOLOv3 trained with RICAP (bottom row).
Without RICAP, a zebra hidden in a tree was detected as two different objects in the left image, side-by-side buses were not detected in the middle image, and a horse tail was misdetected as the dog in the right image, respectively.
RICAP solved these issues, indicating that RICAP makes YOLOv3 be robust to the occlusion.
}
\label{fig:detection}
\end{figure*}

\indent
\subsection{Evaluation on Object Detection}
\label{sec:detection}
\paragraph*{\textbf{Object Detection}}
In this section, we evaluate RICAP on an object detection task.
The main goal is to detect the objects and their position from a given image.
A training image depicts multiple objects with their class information and bounding boxes.
Each bounding box consists of center coordinates, width and height.
Models have to learn and inference these information; object detection is more complicated than image-level classification.
The performance is often evaluated using mean average precision (mAP) and inference time.
YOLO~\cite{Redmon2016} is an end-to-end model achieving faster inference than previous methods.
YOLO was updated to version $3$ (YOLOv3)~\cite{Redmon2018} by the original authors, and we used it as a baseline of our experiments.
\paragraph*{\textbf{Application RICAP to YOLOv3}}
In the object detection, we cannot mix the bounding box labels even if the object is cropped and patched because they are learned using mean squared loss.
Hence, in this case, we only performed the random cropping and patching for input image.
When the object is cropped, we corrected the coordinates, width, and height of bounding box based on the cropped region.
By this change, models cannot obtain the benefit of soft labels, but they can learn and detect of partial features thanks to RICAP.
\paragraph*{\textbf{Experiments and Results}}
We used the Microsoft COCO dataset~\cite{Lin2014} for object detection.
$117,248$ images are used for training and $5,000$ for test.
Each object in image is manually aligned the bounding box and assigned one of $80$ class labels.
We resized all training and test images to $416\times416$.

We performed OpenCV-based data augmentation (\url{https://opencv.org/}).
The weight parameters were updated using the Adam optimizer~\cite{Kingma2014} over $100$ epochs with a mini-batch size of $16$.
The learning rate was initialized to $0.001$.
The backbone model was the DarkNet-53 pre-trained on ImageNet, which is used as the basic backbone in the original study.

Table~\ref{tab:detection} summarizes the experimental results.
We evaluated the hyperparameter values $\beta=0.3$ and $1.0$ fo RICAP.
In addition to mAP, we also show precisions and recalls.
The results demonstrate that RICAP improved the detection performance.
Fig.~\ref{fig:detection} demonstrates difference in detection behavior between YOLOv3 without and with RICAP.
Without RICAP, a zebra hidden in a tree was detected as two different objects in the left image, side-by-side buses were not detected in the middle image, and a horse tail was misdetected as the dog in the right image, respectively.
RICAP solved these issues as shown in images in the bottom row, indicating that RICAP makes YOLOv3 be robust to the occlusion.

\begin{table}[!t]
\vspace*{5mm}
\renewcommand{\arraystretch}{1.5}
\caption{Results of Object Detection using Microsoft COCO.}
\label{tab:detection}
\centering
  \begin{tabular}{lllll}
    \toprule
      Model                         & Precision & Recall & mAP \\
    \midrule
      YOLOv3                        & 54.7      & 52.7   & 51.3 \\
    \midrule
      YOLOv3 + RICAP ($\beta=0.3$)  & \textbf{56.3}      & \textbf{54.1}   & \textbf{52.7} \\
      YOLOv3 + RICAP ($\beta=1.0$)  & \textbf{55.7}      & \textbf{53.5}   & \textbf{52.2} \\
    \bottomrule
  \end{tabular}
\vspace*{5mm}
\end{table}


\section{Conclusion}
In this study, we proposed a novel data augmentation method called \emph{random image cropping and patching} (\emph{RICAP}) to improve the accuracy of the image classification.
RICAP selects four training images randomly, crops them randomly, and patches them to construct a new training image.
Experimental results demonstrated that RICAP improves the classification accuracy of various network architectures for various datasets by increasing the variety of training images and preventing overfitting.
The visualization results demonstrated that RICAP prevents deep CNNs from overfitting to the most apparent features.
The results of the image-caption retrieval task demonstrated that RICAP is applicable to image processing tasks other than classification.

\section*{Acknowledgment}
This study was partially supported by the MIC/SCOPE \#172107101.


\renewcommand\lstlistingname{Algorithm}
\begin{table*}[h]
\begin{lstlisting}[language=Python,caption=Python code for RICAP,basicstyle=\small,frame=lines,label={ricap_algorithm1}]
beta = 0.3 # hyperparameter
for (images, targets) in loader:

    # get the image size
    I_x, I_y = images.size()[2:]

    # draw a boundary position (w, h)
    w = int(numpy.round(I_x * numpy.random.beta(beta, beta)))
    h = int(numpy.round(I_y * numpy.random.beta(beta, beta)))
    w_ = [w, I_x - w, w, I_x - w]
    h_ = [h, h, I_y - h, I_y - h]

    # select and crop four images
    cropped_images = {}
    c_ = {}
    W_ = {}
    for k in range(4):
        index = torch.randperm(images.size(0))
        x_k = numpy.random.randint(0, I_x - w_[k] + 1)
        y_k = numpy.random.randint(0, I_y - h_[k] + 1)
        cropped_images[k] = images[index][:, :, x_k:x_k + w_[k], y_k:y_k + h_[k]]
        c_[k] = targets[index]
        W_[k] = w_[k] * h_[k] / (I_x * I_y)

    # patch cropped images
    patched_images = torch.cat(
        (torch.cat((cropped_images[0], cropped_images[1]), 2)
         torch.cat((cropped_images[2], cropped_images[3]), 2)),
        3)

    # get output
    outputs = model(patched_images)

    # calculate loss
    loss = sum([W_[k] * F.cross_entropy(outputs, c_[k]) for k in range(4)])

    # optimize
    ...
\end{lstlisting}
\end{table*}

\begin{IEEEbiography}[{\includegraphics[width=1in,height=1.25in,bb= 0 0 596 842,clip,keepaspectratio]{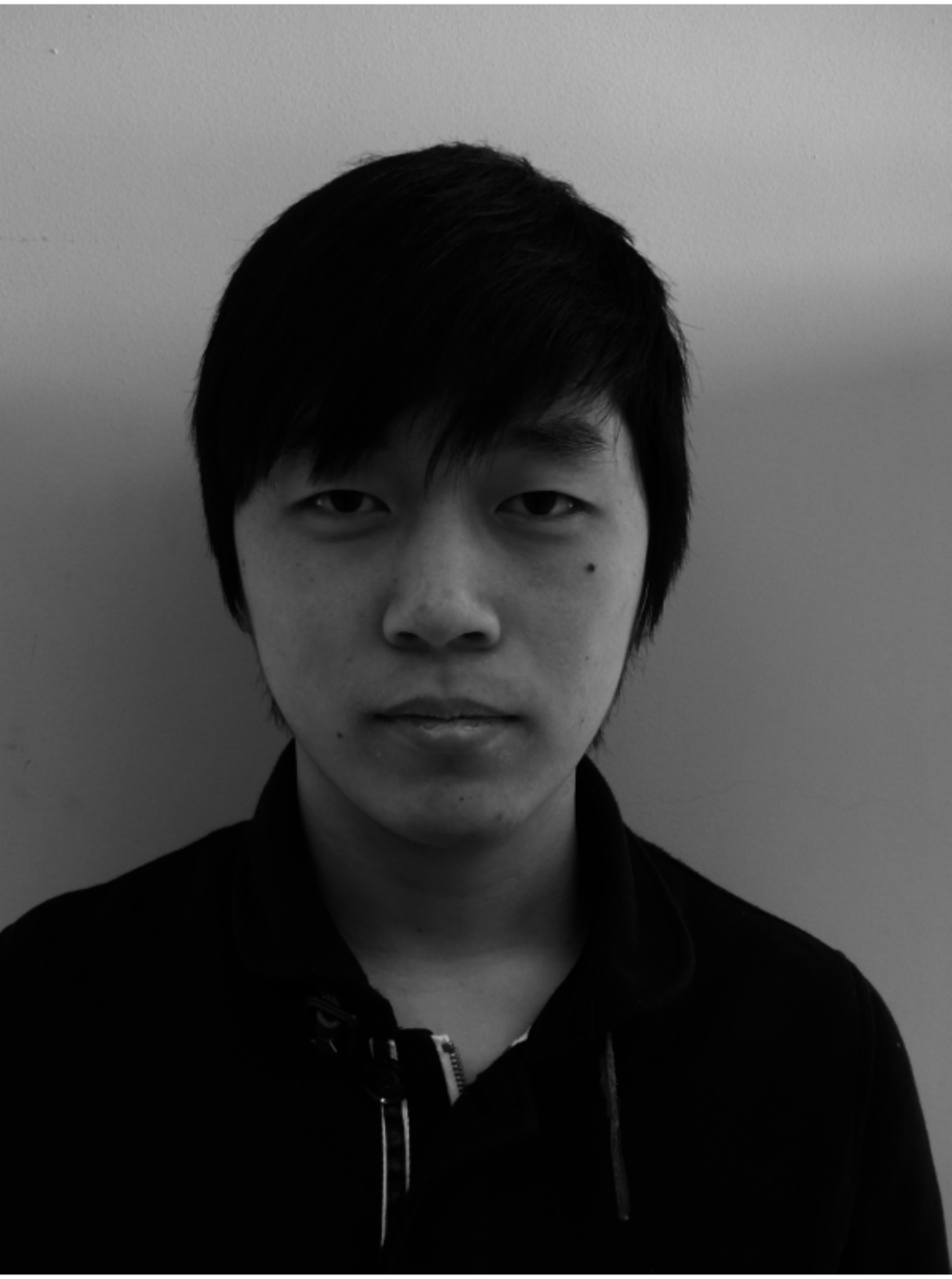}}]%
{Ryo Takahashi}
is a graduate student at the Graduate School of System Informatics, Kobe University, Hyogo, Japan. He received his B.E.~degree from Kobe University. He investigated image classification using deep neural network architectures.
\end{IEEEbiography}

\begin{IEEEbiography}[{\includegraphics[width=1in,height=1.25in,bb= 0 0 360 480,clip,keepaspectratio]{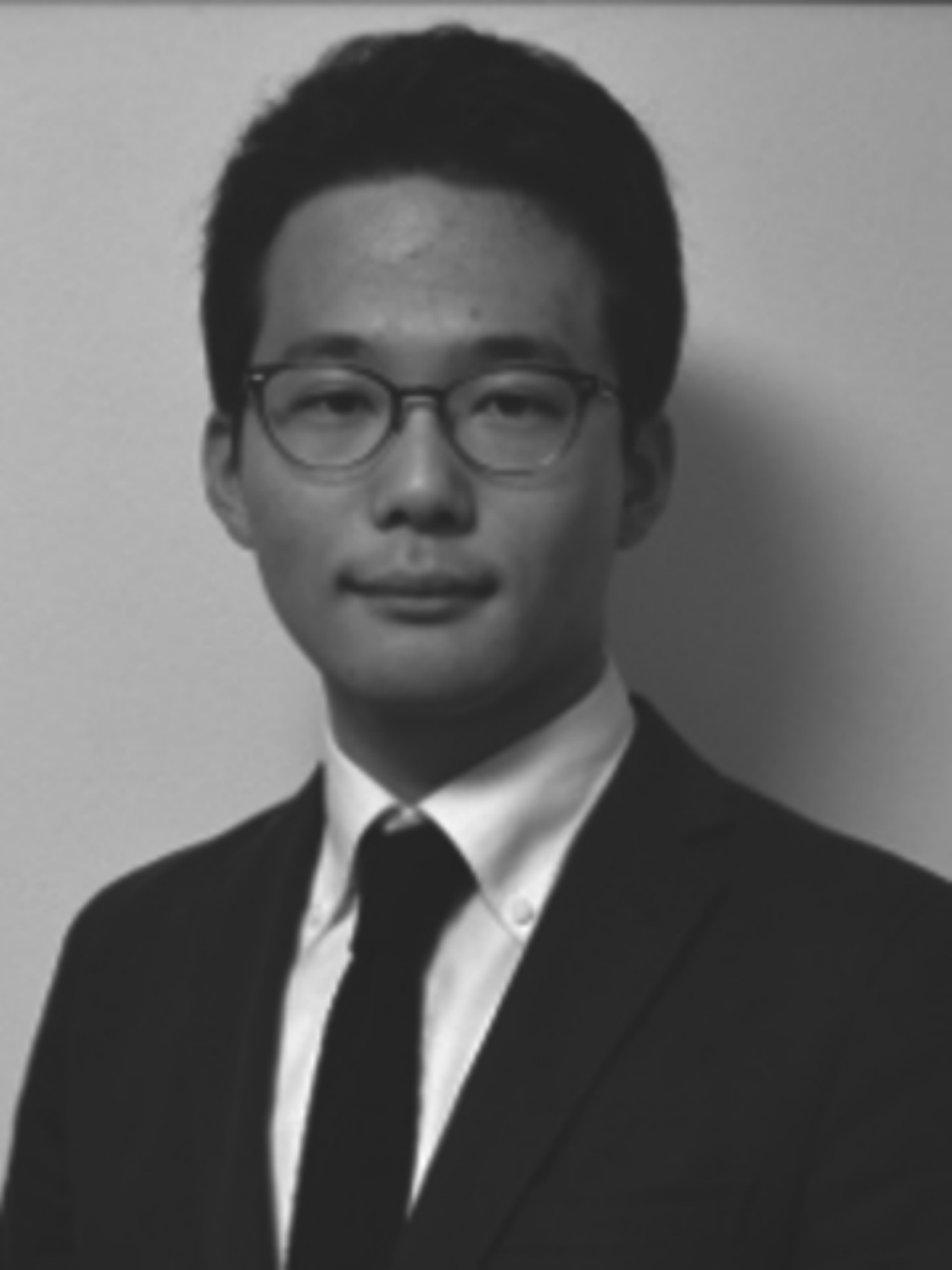}}]%
{Takashi Matsubara}
received his B.E., M.E., and Ph.D.~degrees in engineering from Osaka University, Osaka, Japan, in 2011, 2013, and 2015, respectively.
He is currently an Assistant Professor at the Graduate School of System Informatics, Kobe University, Hyogo, Japan. His research interests are in computational intelligence and computational neuroscience.
\end{IEEEbiography}

\begin{IEEEbiography}[{\includegraphics[width=1in,height=1.25in,bb=0 0 162 180,clip,keepaspectratio]{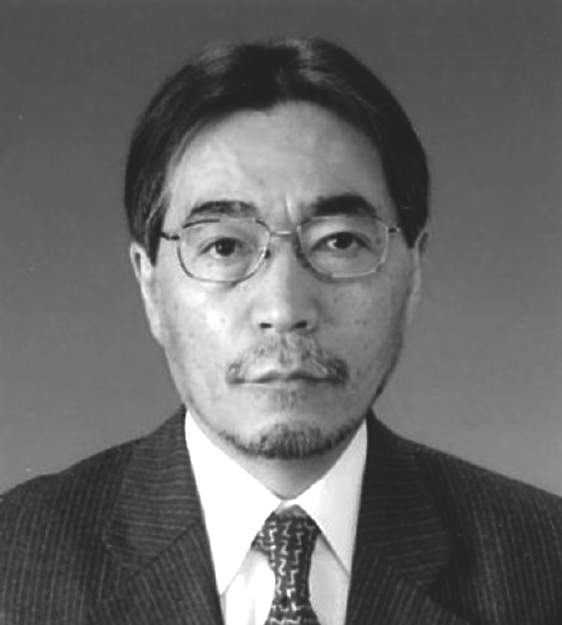}}]%
{Kuniaki Uehara}
received his B.E., M.E., and D.E.~degrees in information and computer sciences from Osaka University, Osaka, Japan, in 1978, 1980 and 1984, respectively. From 1984 to 1990, he was with the Institute of Scientific and Industrial Research, Osaka University as an Assistant Professor. From 1990 to 1997, he was an Associate Professor with the Department of Computer and Systems Engineering at Kobe University. From 1997 to 2002, he was a Professor with the Research Center for Urban Safety and Security at Kobe University. Currently, he is a Professor with the Graduate School of System Informatics at Kobe University.
\end{IEEEbiography}

\section*{Appendix}

\subsection{Code}
For reproduction, we provide a Python code of RICAP in Algorithm~\ref{ricap_algorithm1}.
The code uses numpy and PyTorch (torch in code) modules and follows a naming convention used in official PyTorch examples.
Also, we released the executable code for classification using WideResNet on \url{https://github.com/jackryo/ricap}.

\end{document}